\documentclass[journal]{IEEEtran}

\usepackage[T1]{fontenc}
\usepackage[utf8]{inputenc}

\usepackage{amsmath,amssymb,amsfonts}
\usepackage{bm}
\usepackage{graphicx}
\usepackage{booktabs}
\usepackage{array}
\usepackage{multirow}
\usepackage{makecell}
\usepackage{tabularx}
\usepackage{url}
\usepackage{xcolor}
\usepackage{listings}
\usepackage{fancyvrb}
\usepackage{enumitem}
\usepackage{placeins}
\usepackage{cite}

\setlist[itemize]{leftmargin=2em,topsep=0.3em,itemsep=0.2em,parsep=0pt}
\setlist[enumerate]{leftmargin=2em,topsep=0.3em,itemsep=0.2em,parsep=0pt}

\DeclareMathOperator*{\BCE}{BCE}

\usepackage{hyperref}
\hypersetup{
  colorlinks=true,
  linkcolor=blue!60!black,
  citecolor=blue!60!black,
  urlcolor=blue!60!black,
}

\begin{document}

\title{OctMesh: A Unified Octree-Hierarchical Framework for Lossless Triangle Mesh Compression}

\author{Shiyu~Feng, Xihua~Sheng, Lingyu~Zhu, Chunyang~Fu, and~Shiqi~Wang%
\thanks{The authors are with the Department of Computer Science, City University of Hong Kong, Hong Kong SAR, China (e-mail: shiyufeng9-c@my.cityu.edu.hk; xihsheng@cityu.edu.hk; lingyzhu@cityu.edu.hk; chunyanfu2-c@my.cityu.edu.hk; shiqwang@cityu.edu.hk).}}

\IEEEtitleabstractindextext{%
\begin{abstract}
Lossless triangle mesh compression must preserve both vertex positions and connectivity. Octrees support learned point cloud geometry coding and progressive refinement, but extending them to meshes requires a compatible connectivity representation. Unlike the eight occupancy decisions of a voxel, a parent edge can develop into varied child connections, making edge refinement difficult to model with a compact prediction prior. We propose OctMesh, a learned framework that codes geometry and connectivity on a shared octree hierarchy. Its key observation is that octree pooling produces parents with either one child or two to eight children. Child edges are then grouped by their endpoint parents' types and whether the endpoints share a parent. Each candidate group contains children from just one parent or two connected parents. The resulting four categories define small, fixed-shape prediction tasks: connections uniquely determined by the parent graph are inherited without bits, while three neural predictors estimate probabilities for the remaining candidates. These probabilities guide arithmetic coding of the actual edge symbols. Binarized predictions of within-parent connections provide context for predicting connections between different parents. A graph-aware parent feature extractor combines local geometry, parent connectivity and global shape. The connectivity models use dedicated weights at coarse levels and share weights at fine levels. Residual edges and a finest-level face-selection payload complete the reconstruction. On 256 frames from eight MPEG V-DMC test sequences, OctMesh losslessly recovers the finest-level vertex-coordinate, edge and unoriented face sets at an average of 7.033 bits per face, $12.8\%$ below V-Mesh. The same hierarchical representation supports nine levels of progressive vertex-and-edge refinement.
\end{abstract}

}

\maketitle
\IEEEdisplaynontitleabstractindextext
\IEEEpeerreviewmaketitle

\section{Introduction}

Triangle meshes are widely used in industrial CAD, digital twins, volumetric video, AR/VR, film visual effects and digital heritage. They describe geometry through vertex coordinates and connectivity through faces and edges. Higher mesh resolution and longer sequences increase storage and transmission costs. Lossless compression preserves both types of information exactly, supporting industrial archiving, medical and heritage data, and downstream tasks that require exact reconstruction.

Existing mesh compression methods can be grouped by coding order into connectivity-first and geometry-first approaches. Connectivity-first methods encode adjacency through a deterministic traversal, then predict geometry from the decoded connectivity. Edgebreaker~\cite{rossignac1999} traverses adjacent triangles to produce a compact connectivity symbol stream. Valence-based methods also exploit vertex degree statistics~\cite{touma1998,alliez2001}. Draco~\cite{draco} and the base-mesh codec of MPEG V-DMC (V-Mesh)~\cite{iso23090} use this traversal-based paradigm. Their traversal depends on a fixed set of vertices and faces, making it difficult to extend across resolution levels.

\begin{figure}[!t]
\centering
\includegraphics[width=\columnwidth]{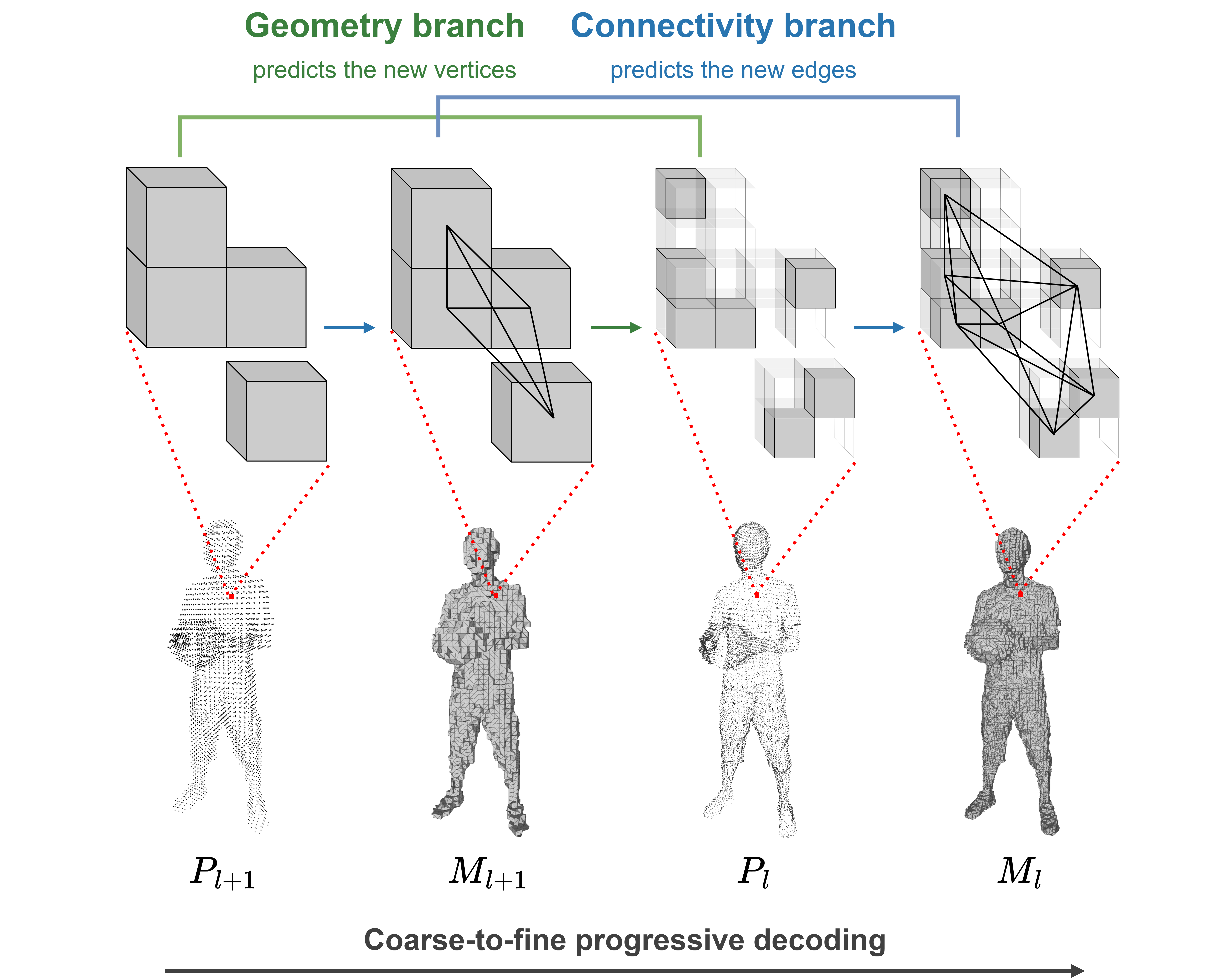}
\caption{Mesh compression is organized on a multiscale octree. The geometry branch refines the parent vertices $P_{l+1}$ into the child vertices $P_l$ by predicting octant occupancy, and the connectivity branch grows the parent mesh $M_{l+1}$ into the child mesh $M_l$ by predicting the newly added edges. }
\label{fig:teaser}
\end{figure}

Geometry-first methods encode vertex coordinates before recovering connectivity on the decoded geometry. Distance-ranked~\cite{marais2007distance} and priority-based~\cite{dvorak2023priority} connectivity coders use known vertex positions to predict triangle connections. A separate multiscale line includes progressive meshes~\cite{hoppe1996}, spectral compression~\cite{karni2000} and progressive geometry compression~\cite{khodakovsky2000}. They use local mesh updates or geometric transforms to organize reconstruction across scales. These input-specific hierarchies require per-vertex operation records or transform coefficients, rather than a shared set of local refinement symbols. Remeshing-based methods also change the original connectivity. DMGC~\cite{zhao2023} combines learned geometry coding with autoregressive connectivity prediction. It scans many candidate vertex pairs without using a multiscale structure to constrain the search. The challenge is to give connectivity refinement a compact symbol structure that a shared neural entropy model can learn.

Learned point cloud compression provides a useful reference. OctSqueeze~\cite{huang2020}, OctAttention~\cite{fu2022}, Unicorn~\cite{wang2024} and SparsePCGC~\cite{wang2023} model occupancy on a coarse-to-fine hierarchy. After octree downsampling, neural models predict each parent voxel's eight-bit occupancy from available context, and binary arithmetic coding transmits the actual bits. Three properties make this approach useful. First, successive levels support progressive level-of-detail (LOD) reconstruction. Second, eight child occupancy bits admit at most 256 patterns, providing a compact discrete prior. Third, sparse convolution~\cite{yan2018,choy2019} and neighborhood attention scale the networks to millions of points.

Extending multiscale neural entropy coding to mesh connectivity remains challenging. Existing attempts either use sparse convolution to encode geometry without explicitly coding the input connectivity~\cite{sirsnc2026}, or attach a neural probability head to a traditional Edgebreaker codec. We argue there are two underlying reasons. First, merging vertices and re-indexing faces across levels changes the mesh traversal on which connectivity-first methods rely. Second, voxel occupancy has a fixed eight-bit representation, whereas edge refinement admits far more varied connection patterns. Extending this framework to meshes therefore requires a structured set of edge-prediction tasks alongside occupancy prediction.

The core idea, illustrated in Fig.~\ref{fig:teaser}, is to group child-level edges by the downsampling behavior of their endpoint parents, giving each group a compact prediction prior. These categories also have natural dependencies, so a cascaded context can reuse information across them. These tasks allow geometry and connectivity to be coded within the same multiscale framework.

Concretely, octree downsampling makes every parent vertex either a Displacement parent with exactly one child vertex or a Split parent with 2--8 child vertices. We assign each child-level edge to one of four disjoint categories according to the types of its endpoints' parents and whether the endpoints share a parent. The resulting four-stage pipeline first inherits connections uniquely determined by the parent mesh, then predicts the remaining edges. Predictions of within-parent edges are binarized and reused as context for predicting connections between different parents. Based on this observation, this paper proposes the \textbf{OctMesh} framework, whose main contributions are summarized as follows:
\begin{itemize}
\item \textbf{Unified framework.} To our knowledge, OctMesh is the first end-to-end learned framework for lossless mesh compression on a multiscale octree hierarchy, transferring the multiscale neural entropy models of point cloud compression to meshes and emitting reversible hierarchical bitstreams for both geometry and connectivity.
\item \textbf{Four-stage edge prediction.} We partition child-level edges into four categories by the downsampling behavior of their parent vertices and process them in four stages, combining zero-bit inheritance with neural prediction. A cascaded context mechanism reuses binarized within-parent edge predictions to condition the prediction of connections between different parents.
\item \textbf{Graph-aware and level-aware modeling.} We design GraphAware-PFE (Graph-Aware Parent Feature Extractor), which fuses local geometry, parent-level connectivity and global shape through three parallel attention paths, together with a level-aware design that gives the coarse levels dedicated weights while the fine levels share one set.
\end{itemize}

\textbf{Experimental results.} On the eight MPEG V-DMC sequences, OctMesh losslessly recovers the finest-level vertex, edge and unoriented face sets at an average of 7.033 bits per face (BPF), about $12.8\%$ below V-Mesh. The representation also provides nine levels of vertex-and-edge refinement from $L_1$ to $L_9$, without per-sequence configuration adjustment (see Sec.~\ref{sec:exp}).

An interactive visualization of the octree hierarchy and the four-stage connectivity reconstruction is available online.\footnote{\label{fn:demo}\url{https://hiddengalaxy1.github.io/octmesh-demo/}}

\section{Related Work}

\subsection{Traditional Mesh Compression}
Mesh compression is a long-standing topic in computer graphics~\cite{peng2005survey,maglo2015}, and early work mostly followed a \emph{connectivity-first, geometry-second} order. Deering~\cite{deering1995} first proposed the concept of ``geometry compression,'' organizing meshes into generalized triangle strips for encoding. The topological surgery of Taubin and Rossignac~\cite{taubin1998} cuts the mesh into a vertex spanning tree and a triangle spanning tree to be encoded separately, which became the basis of MPEG-4 3DMC. Connectivity coding then developed along two main lines: one is \emph{valence-based} coding~\cite{touma1998,alliez2001}, which can approach the entropy lower bound of connectivity. The other is \emph{triangle-traversal-based} coding, represented by the cut-border machine~\cite{gumhold1998}, Edgebreaker~\cite{rossignac1999} and Face Fixer~\cite{isenburg2000facefixer}, which grow a region over the mesh and encode connectivity as a symbol sequence (e.g., $\{C,L,E,R,S\}$ in Edgebreaker), while the geometry generally uses parallelogram prediction for residual coding. To handle arbitrary topology uniformly, the TFAN method of Mamou et al.~\cite{mamou2009} uses the triangle fan as the basic unit, supporting both manifold and non-manifold meshes, and was incorporated into the SC3DMC toolset of MPEG-4/AFX. Draco~\cite{draco} integrates Edgebreaker, kd-tree quantization and arithmetic coding. The latest MPEG V-DMC standard~\cite{iso23090,mammou2022vdmc} uses such codecs to encode the connectivity of its base mesh, adding subdivision and displacement coding to support mesh sequences.

With vertex positions already known, distance-ranked coding~\cite{marais2007distance} encodes the rank of each triangle's closing vertex, while priority-based coding~\cite{dvorak2023priority} chooses expansion gates by prediction certainty. Both use hand-crafted geometric predictors.

Progressive methods organize reconstruction across scales. Progressive meshes~\cite{hoppe1996} achieve a coarse-to-fine representation through a sequence of edge-collapse and vertex-split records. Spectral compression~\cite{karni2000} projects geometry onto the eigenbasis of the mesh Laplacian for coding. Progressive geometry compression~\cite{khodakovsky2000} encodes geometry at multiple scales with a wavelet transform after semi-regular remeshing. A line of progressive coders~\cite{pajarola2000,alliez2001progressive,valette2009ipr} extends this idea to lossless transmission by batching vertex splits and predicting them from the decoded coarse mesh. These methods use input-specific refinement operations or geometric transforms; remeshing-based methods also change the original connectivity.

\subsection{Learned Point Cloud Compression}
Learned point cloud geometry compression uses \emph{conditional entropy models over octrees and voxels} to estimate the probabilities of geometry symbols. As in image compression, hyperpriors and decoded context condition symbol probabilities~\cite{balle2018,minnen2018,he2021checkerboard,zhu2024dynamic}. This principle also supports lossless image coding~\cite{mentzer2019,bai2024dlpr}. In parallel, MPEG has advanced two standardization routes, geometry-based G-PCC and video-based V-PCC~\cite{graziosi2020}. Within the learned paradigm, one category models node occupancy directly on the \emph{octree}: OctSqueeze~\cite{huang2020} uses an MLP to predict the conditional probability of the 8-bit occupancy of each octree node, VoxelContext-Net~\cite{que2021} introduces neighborhood voxel context, and OctAttention~\cite{fu2022} further aggregates the context of sibling and ancestor nodes via an attention mechanism, whose cost OctFormer~\cite{cui2023octformer} lowers with windowed attention over node sequences.

Another category works in the \emph{sparse-voxel} domain: VoxelDNN~\cite{nguyen2021} performs autoregressive occupancy modeling with masked 3D convolutions, PCGCv2~\cite{wang2021pcgcv2} and SparsePCGC~\cite{wang2023} realize multiscale occupancy coding via sparse convolution~\cite{yan2018,choy2019}, and Unicorn~\cite{wang2024} proposes a unified multiscale conditional coding framework that achieves leading efficiency across lossy and lossless settings. Related ideas also appear in learned attribute coding~\cite{sheng2022deeppcac,fu2026dald}, and human body priors have been exploited for human point clouds~\cite{wu2024geoprior}. Their bitstreams encode point positions or attributes, without explicit mesh connectivity.

\subsection{Learned Mesh Compression}
Work that directly applies deep learning to mesh compression is still limited. DMGC~\cite{zhao2023} is a representative learned method for \emph{lossless} mesh compression. It follows the geometry-first order: it first encodes geometry with a point cloud codec, then has a network predict the connections between vertex pairs on top of the decoded geometry. Related autoregressive mesh generators use different sequence representations: PolyGen~\cite{polygen} predicts vertices followed by faces, MeshGPT~\cite{meshgpt} uses learned mesh tokens, and MeshXL~\cite{meshxl} models coordinate sequences. These methods generate meshes, whereas a codec represents a given input mesh.

A larger body of recent methods targets \emph{lossy} mesh compression through learned surface representations. NGF~\cite{ngf2024} partitions the surface into quadrilateral patches and optimizes per-patch feature vectors together with a shared coordinate network under a differentiable-rendering loss. NDF~\cite{ndf2025} encodes a quantized neural displacement field that deforms a subdivided coarse template toward the input surface. SIR-SNC~\cite{sirsnc2026} voxelizes the input into a near-surface sparse TSDF and compresses it with a sparse-convolutional autoencoder plus a learned entropy model, extracting an iso-surface at the decoder. Their reconstructed connectivity need not match that of the input mesh. Neural Progressive Meshes~\cite{chen2023npm} transmit learned residual features across subdivision levels to progressively improve surface reconstruction. They operate on a remeshed subdivision hierarchy and target an approximation of the input surface. Our focus is learned geometry and edge coding on a shared octree hierarchy, with progressive refinement and finest-level face recovery.

\section{Method}

\begin{figure*}[t]
\centering
\includegraphics[width=\textwidth]{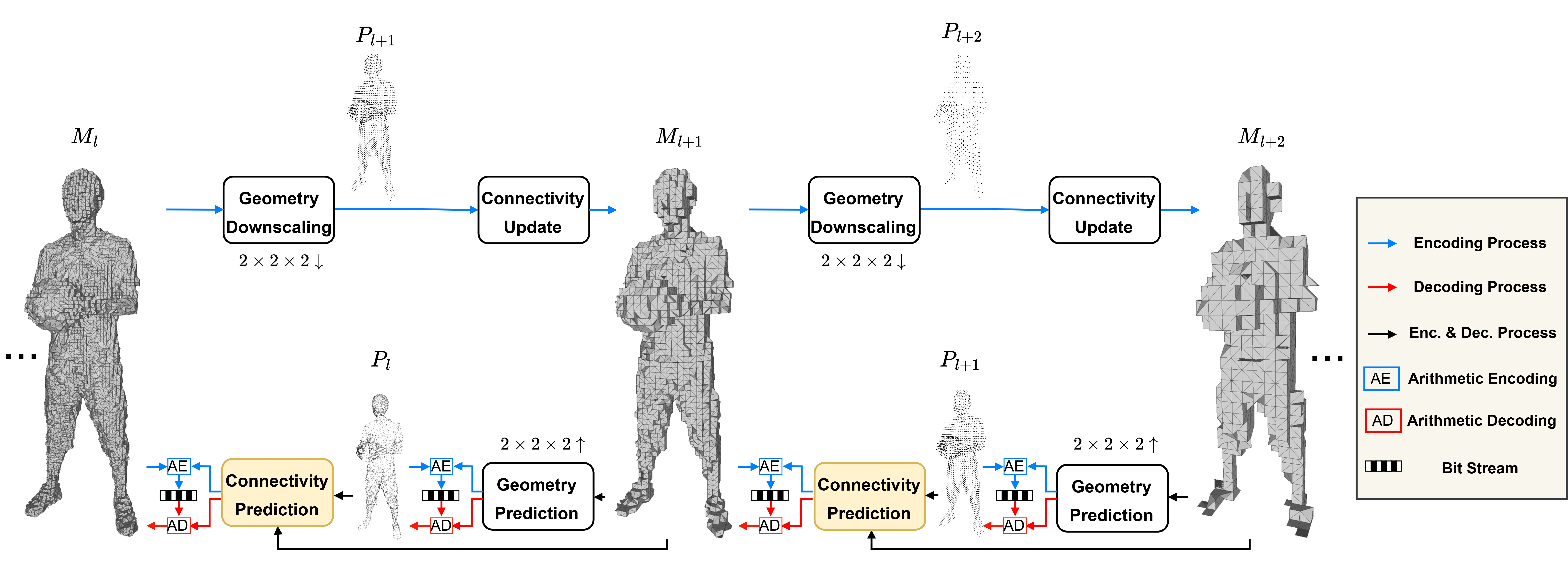}
\caption{Overall OctMesh framework. Top, left to right: octree pooling ($2\!\times\!2\!\times\!2\,\downarrow$) and connectivity updates build the hierarchy from fine to coarse. Bottom, right to left: the geometry branch first recovers child vertices, and the connectivity branch then recovers their edges using the parent mesh. Both branches predict probabilities for arithmetic encoding (AE) and decoding (AD). Blue and red arrows indicate encoding and decoding, respectively; black arrows indicate operations shared by both.}
\label{fig:overview}
\end{figure*}

\begin{table}[!t]
\centering
\caption{Summary of Main Symbols}
\label{tab:notation}
\renewcommand{\arraystretch}{1.15}
\begin{tabularx}{\columnwidth}{@{}l@{\hspace{0.7em}}>{\raggedright\arraybackslash}X@{}}
\toprule
Symbol & Meaning \\
\midrule
\makecell[l]{$M_l=(P_l,F_l,E_l)$\\$l=1,\ldots,L$} & Mesh vertices, faces, edges; $L$ levels ($1$ finest, $L$ base) \\
$p$, $f$, $e$ & Vertex; triangular face $f=(p_1,p_2,p_3)$; undirected edge \\
\makecell[l]{$p'=\pi(p)$\\$\mathrm{ch}(p')$} & Parent mapping; children of parent $p'$ \\
$\tau(p')$ & Parent type: Disp (one child) or Split (2--8 children) \\
$\mathrm{oct}(p)$, $\mathbf{o}^{(p')}$ & Octant index (1--8); 8-entry binary occupancy \\
\makecell[l]{$\mathbf{g}^{(p')}$\\$d$, $d'=d/2$} & GraphAware-PFE feature; feature width $d$; bilinear output width $d'$ \\
$\mathbf{A}^{(p')}$, $\mathbf{N}$ & Binary Split-Intra and Split-Split target matrices \\
$\hat{a}^{(p')}_{ij}$, $\tilde{\mathbf{A}}^{(p')}$ & Split-Intra probability; binarized prediction matrix \\
$S$, $S_a$, $S_b$ & Per-octant internal connectivity context in figures \\
\makecell[l]{$\mathrm{DD}$, $\mathrm{SI}$, $\mathrm{SD}$, $\mathrm{SS}$\\$E_l^{\mathrm{DD}}$} & Disp-Disp, Split-Intra, Split-Disp, Split-Split; inherited DD edges \\
$N_{l+1}$, $K$, $G$ & $N_{l+1}=|P_{l+1}|$ parents; $K$ neighbors; $G$ global tokens \\
$y_i$, $\hat{y}_i$ & Binary target; probability of symbol 1 \\
\makecell[l]{$\ell_{l,s}$, $\mathcal{T}$\\$\mathcal{L}_{\mathrm{edge}}$} & Mean BCE over valid stage-$s$ candidates at level $l$; contributing levels; edge loss \\
$|P_1|$, $|F_1|$ & Finest-level vertex and face counts \\
$B$, $\mathrm{BPF}=B/|F_1|$ & Total stream bits; bits per finest-level face \\
$D$ & Larger of two directional mean vertex-to-triangle distances \\
\bottomrule
\end{tabularx}
\end{table}

\subsection{Overall Framework}\label{ssec:framework}
OctMesh codes a triangle mesh on a multiscale octree hierarchy with two branches: a \emph{geometry branch} that predicts the vertices of each level and a \emph{connectivity branch} that predicts the edges. The geometry branch reuses the occupancy-coding paradigm that is already mature in point cloud compression, and we leave it unchanged; the rest of this section therefore focuses on the connectivity branch. We first fix the notation, which is used throughout this section and summarized in Table~\ref{tab:notation}. A triangle mesh at level $l$ is written as $M_l=(P_l,F_l,E_l)$, where $P_l$ is the vertex set, $F_l$ the face set and $E_l$ the edge set. Each vertex $p\in P_l$ is an integer coordinate in $\mathbb{Z}^3$, each face $f=(p_1,p_2,p_3)\in F_l$ is a triple of vertices, and each edge $e\in E_l$ is an unordered vertex pair taken from the sides of the triangles, so $E_l$ is fully determined by $F_l$. The input mesh is \emph{voxel-aligned}: each occupied voxel contains exactly one vertex. The preprocessing that yields such an input is described in Sec.~\ref{sec:impl}.

OctMesh organizes the input into an $L$-level hierarchy $\{M_l=(P_l,F_l,E_l)\}_{l=1}^{L}$ by octree pooling (Sec.~\ref{ssec:pooling}). The finest level $M_1$ is the input mesh, and the coarsest level $M_L$ is the base mesh. The mesh sets $P_l$, $F_l$ and $E_l$ carry the level subscript $l$, while individual elements are written as a vertex $p$, a face $f$ or an edge $e$. During one level transition, we call a vertex of the finer level $P_l$ a \emph{child vertex} and the corresponding vertex $p'$ of the coarser level $P_{l+1}$ its \emph{parent vertex}; two child vertices are written $p_a$, $p_b$ and two parent vertices $p'_a$, $p'_b$. Our goal is lossless coding of the vertex coordinates, edges and unoriented faces of the finest mesh $M_1$.

Figure~\ref{fig:overview} shows the overall coding flow. Before encoding, the hierarchy is built from fine to coarse by octree pooling and connectivity update, detailed in Sec.~\ref{ssec:pooling}. Decoding proceeds in the opposite direction, from coarse to fine. For each transition $M_{l+1}\to M_l$, the geometry branch first predicts the occupancy probability of every candidate child vertex, and the connectivity branch then predicts the existence probability of every candidate child edge. Under these probabilities, binary arithmetic encoding (AE)~\cite{witten1987} writes the actual bits into the bitstream. The decoder runs the same networks, reproduces exactly the same probabilities, and reads the bits back by arithmetic decoding (AD). This process refines the decoded vertices and edges level by level. The bitstream stores the base-mesh record and the per-level geometry and connectivity records, which the decoder accesses in coarse-to-fine order.

\subsection{Octree Pooling and Connectivity Update}\label{ssec:pooling}

\begin{figure*}[t]
\centering
\includegraphics[width=\textwidth]{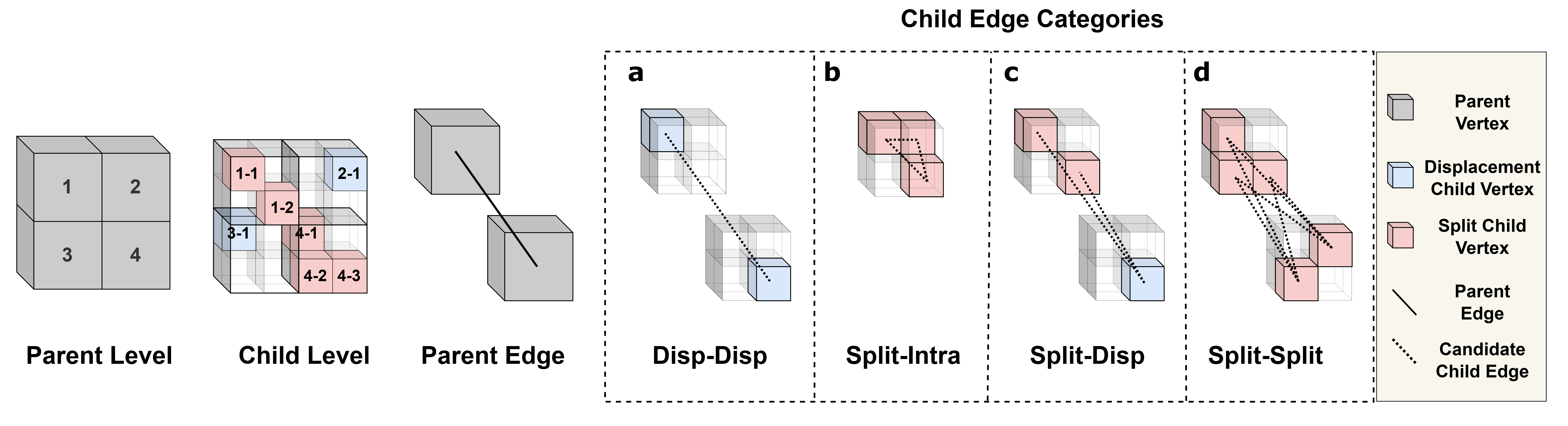}
\caption{Four-category partition of child-level edges. Left: parent vertices 1 and 4 are Split parents, and 2 and 3 are Displacement parents. Light-blue and pink octants contain Displacement and Split children, respectively. Right: (a) Disp-Disp, (b) Split-Intra, (c) Split-Disp and (d) Split-Split distinguish whether the endpoints share a parent and the types of their parents. Black solid lines represent parent edges; black dotted lines represent candidate child edges.}
\label{fig:edge-cats}
\end{figure*}

We now describe how the hierarchy of Sec.~\ref{ssec:framework} is built, level by level from fine to coarse.

\textbf{Octree pooling.} Downsampling between adjacent levels is a stride-2 voxel pooling. Each vertex $p\in P_l$ is mapped to the coarser coordinate $\lfloor p/2 \rfloor$, and the vertices falling into the same coarse voxel are merged into one parent vertex. Deduplicating these coordinates gives the parent vertex set $P_{l+1}$. The mapping $\pi:P_l\to P_{l+1}$ records the parent $p'=\pi(p)$ of every vertex, and the preimage $\mathrm{ch}(p')=\{p:\pi(p)=p'\}$ collects the child vertices of every parent vertex. Since a coarse voxel covers a $2\!\times\!2\!\times\!2$ block of fine voxels, each child vertex has exactly one parent vertex and each parent vertex has 1 to 8 child vertices.

\textbf{Connectivity update.} Merging the vertices alone does not give a valid mesh, so the faces must be updated as well. Each face $f=(p_1,p_2,p_3)\in F_l$ is re-mapped to the parent level as $f'=\big(\pi(p_1),\pi(p_2),\pi(p_3)\big)$. When two or all three of its vertices land in the same parent vertex, $f'$ collapses and is discarded; the remaining faces are deduplicated to form the parent face set $F_{l+1}$. The remaining triangles then determine the parent edge set $E_{l+1}$, whose edges are simply the triangle sides. The resulting level is $M_{l+1}=(P_{l+1},F_{l+1},E_{l+1})$, with repeated-index faces and duplicate faces removed.

\subsection{Parent Types and Edge Categories}\label{ssec:parent-types}\label{ssec:edge-cls}
At each refinement step, the occupancy decoded by the geometry branch tells us how many children each parent has. This count determines the parent type $\tau(p')\in\{\mathrm{Disp},\mathrm{Split}\}$. A \emph{Displacement parent} has one child ($|\mathrm{ch}(p')|=1$), while a \emph{Split parent} has two or more ($|\mathrm{ch}(p')|\geq 2$). Their children are called Displacement and Split children, respectively. We classify child-level edges by their endpoints' parent types and whether they share a parent (Fig.~\ref{fig:edge-cats}). \emph{Disp-Disp} connects children of two Displacement parents. \emph{Split-Intra} connects children of the same Split parent. \emph{Split-Disp} connects a Split child to a Displacement child. \emph{Split-Split} connects children of two different Split parents. Every child-level edge belongs to exactly one category.

To represent the children consistently across these categories, we index each child by its position in the parent's $2\!\times\!2\!\times\!2$ voxel block. For a child at $p=(p_x,p_y,p_z)$, the local octant index is
\begin{equation}
\begin{aligned}
\mathrm{oct}(p) &= 1 + (p_x \bmod 2) + (p_y \bmod 2)\cdot 2 \\
&\quad + (p_z \bmod 2)\cdot 4 \;\in\; \{1,2,\dots,8\}.
\end{aligned}
\end{equation}
The occupancy vector $\mathbf{o}^{(p')}\in\{0,1\}^8$ records this child layout: entry $i$ is 1 if octant $i$ contains a child and 0 otherwise. All parents use the same local numbering. Thus, the numbers 1--8 in Figs.~\ref{fig:stage21}--\ref{fig:stage23} identify octants within the parent represented by each cube or matrix axis.

EdgePartitioner uses these occupied octants to form candidate child edges. Within a Split parent, every pair of distinct children is a candidate. For each parent edge, we also pair every child of one endpoint with every child of the other. These pairs are possible connections whose existence must be determined during coding. Each candidate group uses children of only one or two parents. With at most eight children per parent, the group remains small even for a large mesh.

\begin{figure}[!t]
\centering
\includegraphics[width=\linewidth]{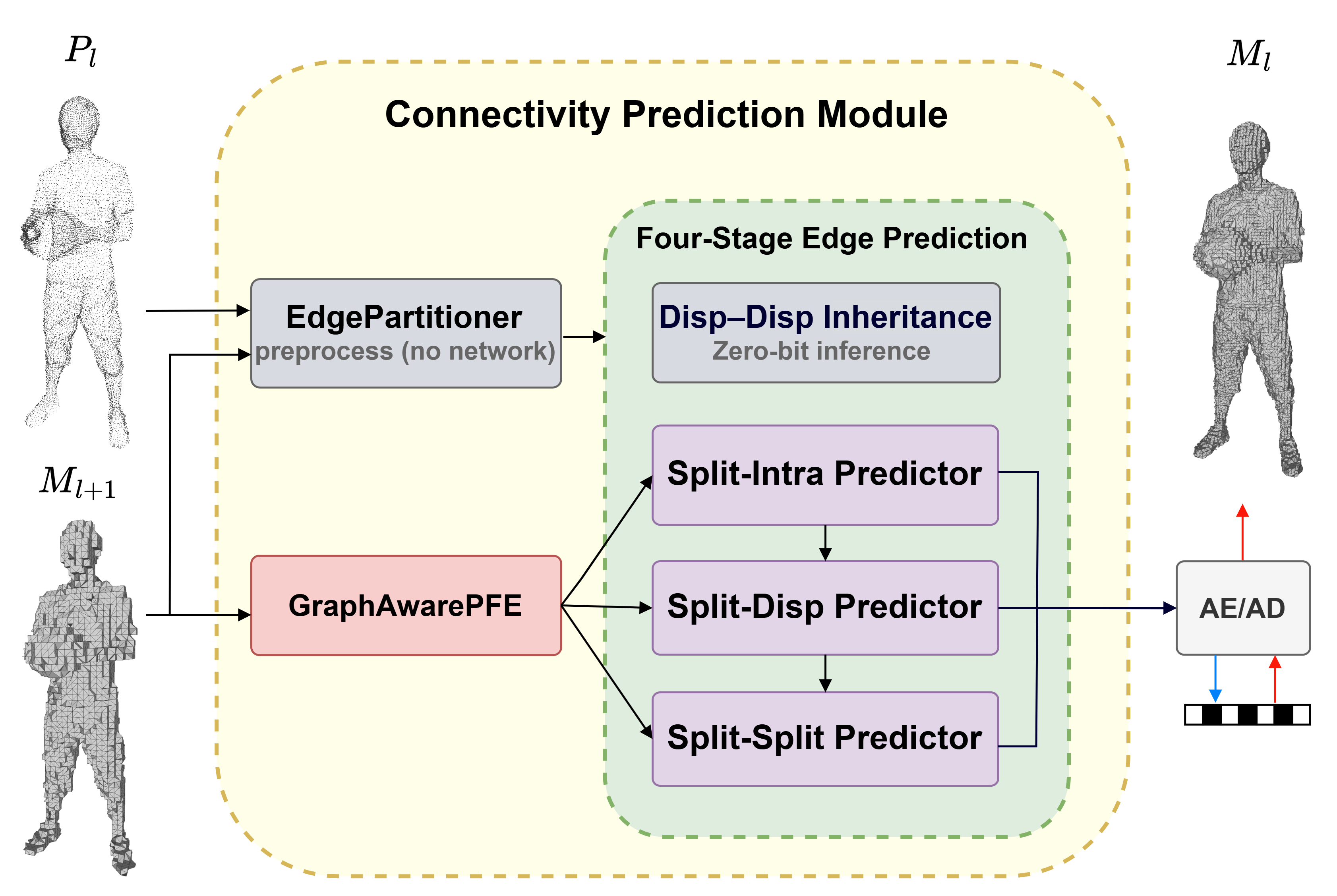}
\caption{Connectivity prediction module. The inputs are the parent mesh $M_{l+1}$ and the decoded child vertices $P_l$. EdgePartitioner forms the candidate edge sets, while GraphAware-PFE extracts parent features for the three neural predictors. Disp-Disp edges are inherited without bits. The binarized Split-Intra output provides context to both Split-Disp and Split-Split. Predicted probabilities guide arithmetic encoding/decoding (AE/AD) of the corresponding edge symbols to recover the connectivity of $M_l$.}
\label{fig:cp-module}
\end{figure}

\begin{figure*}[!t]
\centering
\includegraphics[width=\textwidth]{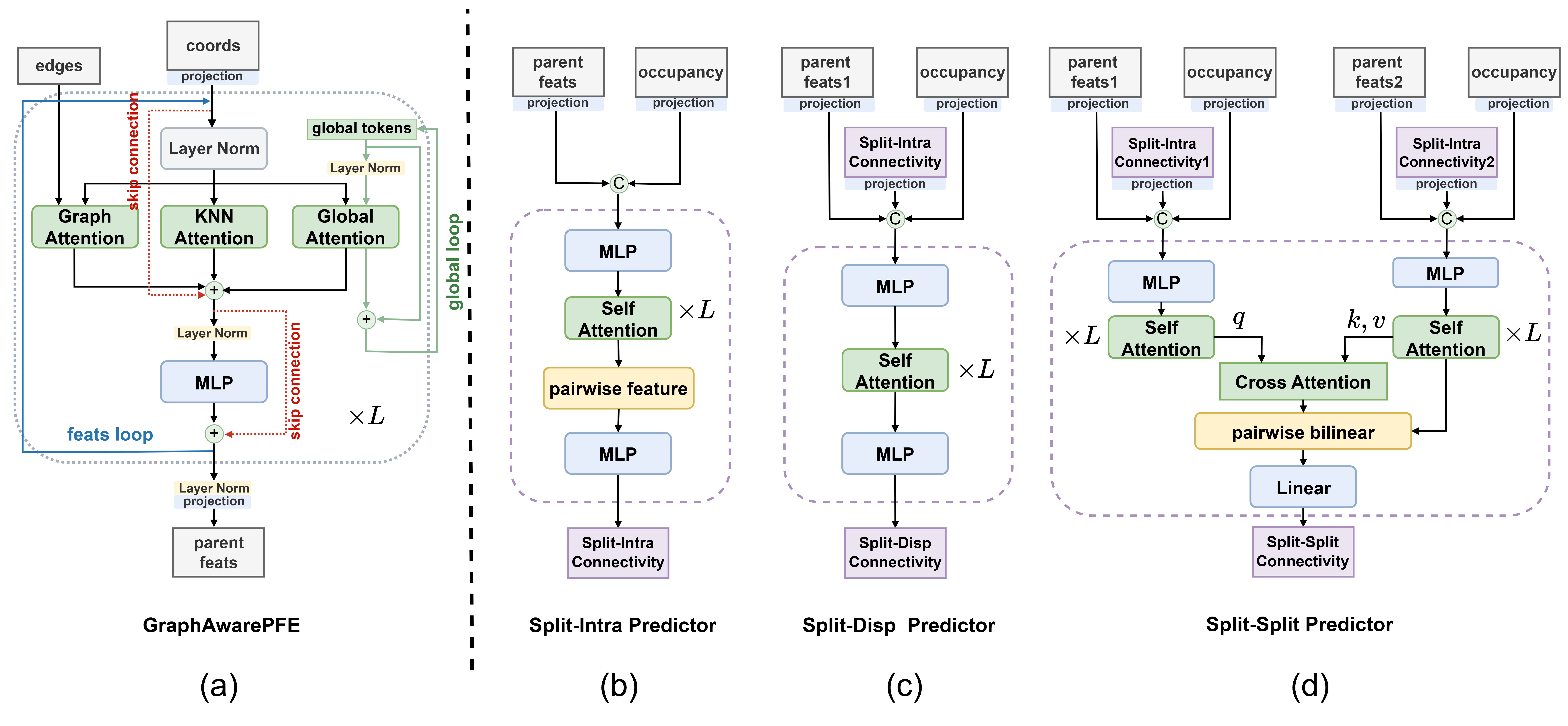}
\caption{Architecture of the learnable modules. (a) GraphAware-PFE combines KNN, graph and global attention to extract parent features. (b) Split-Intra uses octant self-attention and pairwise features to score internal connections. (c) Split-Disp uses the Split parent's internal connectivity context to score connections to a Displacement child. (d) Split-Split uses internal context from both parents, self-attention within each parent, cross-attention between parents and a bilinear head. Circled C and $+$ denote concatenation and addition, respectively.}
\label{fig:netarch}
\end{figure*}

\subsection{Overview of the Connectivity Prediction Module}\label{ssec:cp-overview}
The module in Fig.~\ref{fig:cp-module} first handles the Disp-Disp candidates. For each edge $(p'_a,p'_b)\in E_{l+1}$ between two Displacement parents, there is only one possible child pair $(p_a,p_b)$. Its connection is inherited directly from the parent edge. These connections form the edge set $E_l^{\mathrm{DD}}$ at both the encoder and decoder, without a probability model or additional edge bits.

The remaining categories require learned connection probabilities. To predict them, GraphAware-PFE extracts a feature vector $\mathbf{g}^{(p')}\in\mathbb{R}^d$ for each parent $p'$ from the decoded parent mesh $M_{l+1}$, where $d$ is the feature dimension. This vector summarizes the geometry and connectivity around that parent. The predictors combine these parent features with octant positions and occupancy to describe the children. Split-Intra first estimates connections among children of the same Split parent. These estimates then provide internal connectivity context for both Split-Disp and Split-Split when predicting connections between different parents.

\subsection{Split-Intra Edges}\label{ssec:stage21}

\begin{figure}[!t]
\centering
\includegraphics[width=\linewidth]{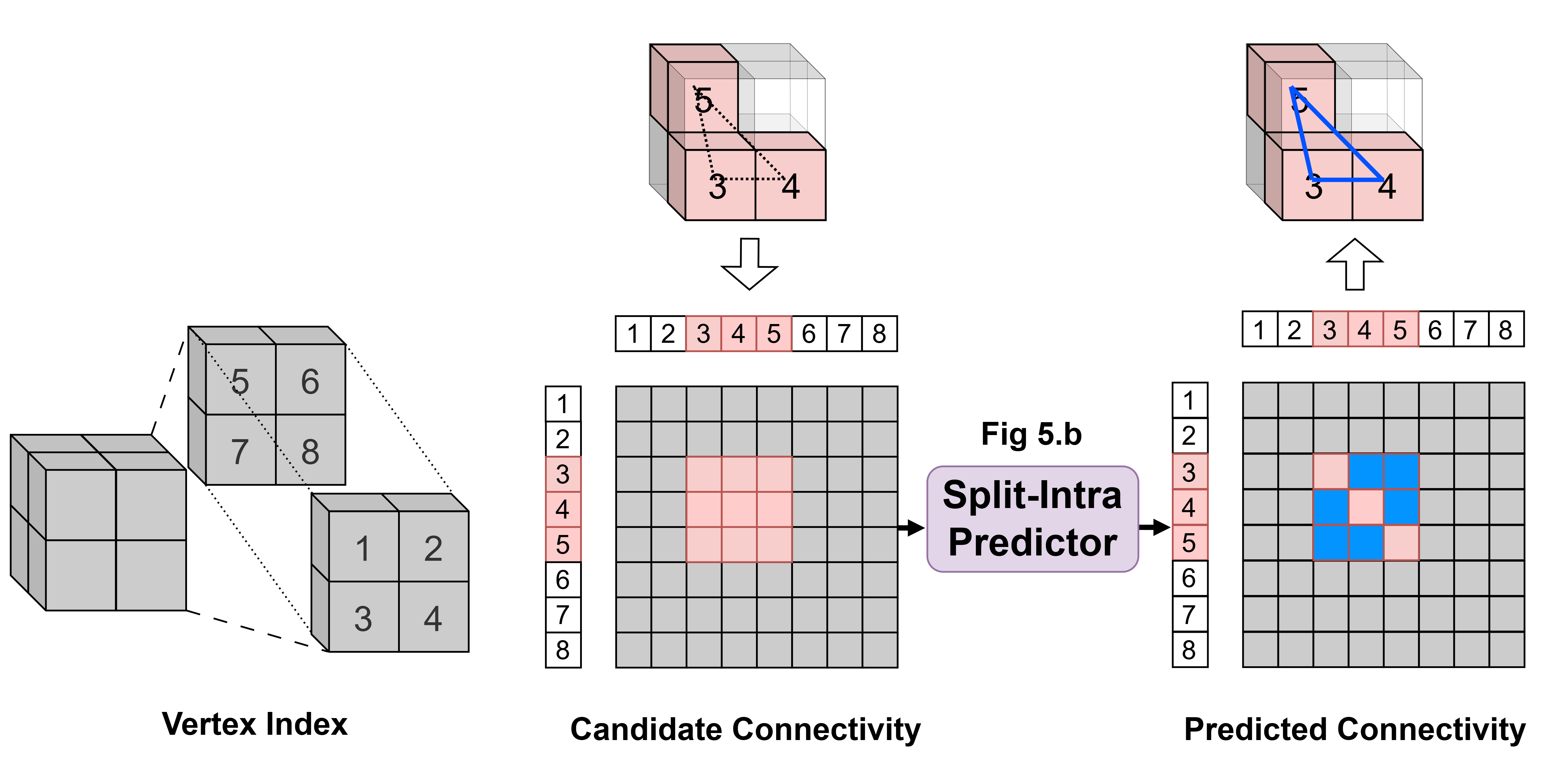}
\caption{Split-Intra connectivity prediction. The left diagram defines the local octant indices 1--8. Both matrix axes use these indices for the same Split parent. The pink block spans occupied octants 3, 4 and 5; gray entries involve empty octants.  On the right, blue cells mark predicted connections after binarization, and blue solid lines show the corresponding edges. }
\label{fig:stage21}
\end{figure}

For a Split parent $p'$, internal connectivity is represented by $\mathbf{A}^{(p')}\in\{0,1\}^{8\times 8}$, with both axes indexing its octants. Entry $\mathbf{A}^{(p')}_{ij}$ is 1 if the children in octants $i$ and $j$ are connected and 0 otherwise. A valid candidate joins two occupied octants with different indices. Since the edges are undirected, the matrix is symmetric and only valid entries in the strict upper triangle are coded. The occupied octants 3, 4 and 5 in Fig.~\ref{fig:stage21} therefore give three binary edge symbols, corresponding to $(3,4)$, $(3,5)$ and $(4,5)$.

For each octant, the Split-Intra predictor in Fig.~\ref{fig:netarch}b combines the shared parent feature $\mathbf{g}^{(p')}$, an encoding of occupancy $\mathbf{o}^{(p')}$ and a positional encoding of that octant's local coordinates. An \emph{Octant Transformer} uses self-attention~\cite{vaswani2017} among occupied octants to produce a feature $\mathbf{z}_i\in\mathbb{R}^d$ for each position. Following Swin Transformer~\cite{liu2021}, a learnable relative position bias also encodes the offset between two octants. Each coordinate offset is $-1$, 0 or 1, giving $3^3=27$ possible offsets and one bias entry for each. Each candidate pair $(i,j)$ is then represented by the endpoint features, their difference and their element-wise product:
\begin{equation}
\mathbf{z}_{ij} = \big[\, \mathbf{z}_i \,\|\, \mathbf{z}_j \,\|\, \mathbf{z}_i - \mathbf{z}_j \,\|\, \mathbf{z}_i \odot \mathbf{z}_j \,\big] \in \mathbb{R}^{4d},
\end{equation}
where $\|$ denotes concatenation and $\odot$ denotes the element-wise product. The difference and product provide pairwise comparisons in addition to the individual endpoint features. A multilayer perceptron (MLP) maps $\mathbf{z}_{ij}$ to a score, or logit. The upper-triangular logits are mirrored to enforce symmetry, and a sigmoid converts them to connection probabilities $\hat{a}^{(p')}_{ij}$.

\subsection{Split-Disp Edges}\label{ssec:stage22}

\begin{figure}[!t]
\centering
\includegraphics[width=\linewidth]{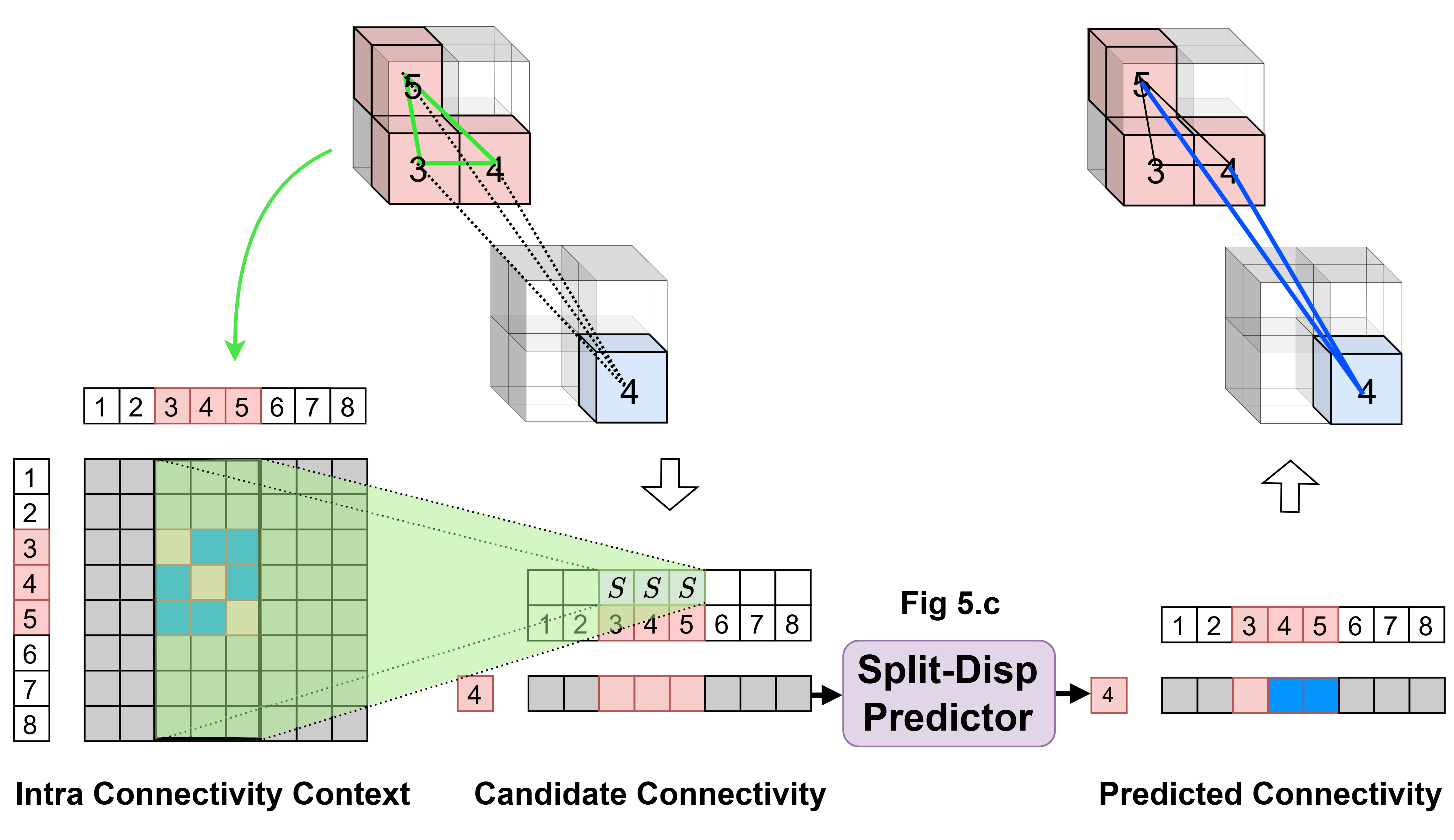}
\caption{Split-Disp connectivity prediction. The row represents the Displacement child in octant 4; pink entries mark candidate connections to Split octants 3, 4 and 5. Each $S$ denotes the corresponding octant's internal connectivity context, illustrated by the green edges within the Split parent. Blue cells and solid blue edges show the binarized predictions.}
\label{fig:stage22}
\end{figure}

For a parent edge joining a Split parent and a Displacement parent, we predict which Split children connect to the unique Displacement child. These candidate connections form a $1\times 8$ row indexed by the Split parent's octants; only occupied columns are valid. Equal octant indices on the two sides refer to distinct vertices: in Fig.~\ref{fig:stage22}, the two children numbered 4 belong to different parents.

For the Split parent $p'$, we convert its previously predicted Split-Intra probabilities $\hat{a}^{(p')}_{ij}$ into the binary matrix $\tilde{\mathbf{A}}^{(p')}$. The column for each occupied octant gives an eight-entry vector of predicted connections to the octants of the same parent. This vector is the context labeled $S$ in Fig.~\ref{fig:stage22}. It describes how that child connects to others within its own group. For example, the vector above octant 3 indicates predicted connections to octants 4 and 5.

The Split-Disp predictor in Fig.~\ref{fig:netarch}c maps each internal connection vector to a learned context feature with an MLP. It combines this feature with the two parent features, octant positional encoding and occupancy. An MLP fuses the inputs, and an Octant Transformer models the relations among the Split children. A final MLP and sigmoid produce the eight connection probabilities.

\subsection{Split-Split Edges}\label{ssec:stage23}

\begin{figure}[!t]
\centering
\includegraphics[width=\linewidth]{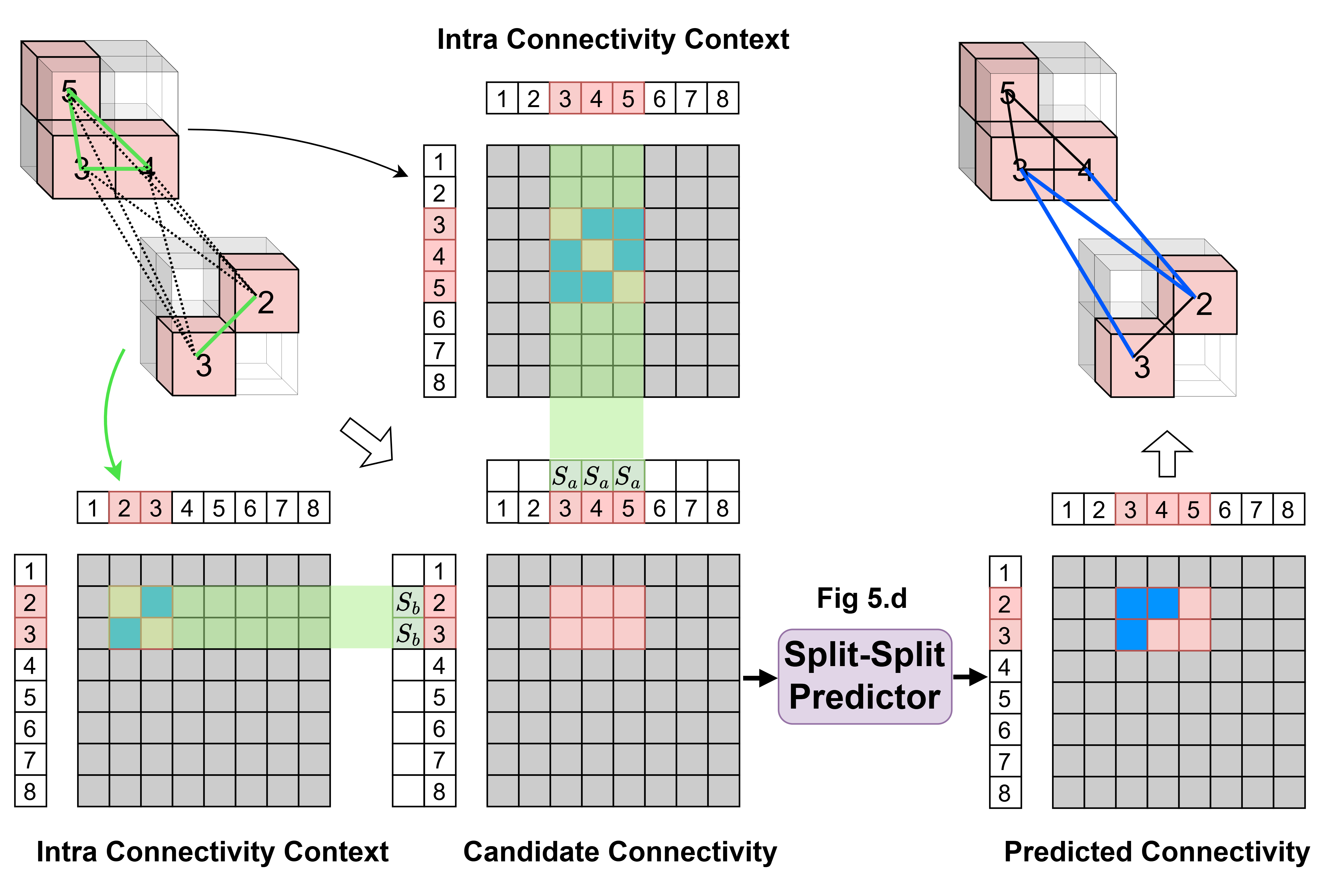}
\caption{Split-Split connectivity prediction. Columns index the upper Split parent's octants and rows index the lower parent's octants. Occupied columns 3, 4 and 5 and rows 2 and 3 define six pink candidates, shown as black dotted edges between the cubes. Green highlights the internal connectivity context: each $S_a$ is an eight-entry column from the upper parent's Split-Intra matrix, and each $S_b$ is an eight-entry row from the lower parent's matrix. On the right, blue cells and solid edges show the three predicted connections after binarization; black solid edges retain the internal connections.}
\label{fig:stage23}
\end{figure}

For a parent edge joining two Split parents, we arrange the candidate connections in an $8\times 8$ matrix $\mathbf{N}$. In Fig.~\ref{fig:stage23}, its rows index the lower cube's octants and its columns index the upper cube's octants. The binary entry $\mathbf{N}_{ij}$ specifies whether the two children at row $i$ and column $j$ are connected. A candidate is valid when both octants are occupied.

The two axes belong to different parents, so $\mathbf{N}_{ij}$ and $\mathbf{N}_{ji}$ generally refer to different vertex pairs. Thus, $\mathbf{N}$ need not be symmetric, even though mesh edges are undirected. Diagonal entries also remain valid because equal octant indices on the two sides refer to distinct children.

As in Split-Disp, each child receives an eight-entry internal connection vector from its own parent's binarized Split-Intra matrix. The labels $S_a$ and $S_b$ in Fig.~\ref{fig:stage23} mark these vectors for the column and row octants, respectively. The Split-Split predictor in Fig.~\ref{fig:netarch}d combines each vector with the corresponding parent feature, positional encoding and occupancy. Self-attention models the relations within each group of children. Cross-attention then uses one group's features as the query and the other's as the key and value, allowing the first group to gather information from the second. The octant relative position bias follows the convention used in Split-Intra.

For each candidate pair, let $\mathbf{a},\mathbf{b}\in\mathbb{R}^{d}$ denote the endpoint features after attention. A bilinear layer combines products of their components with learned weights to form a joint pair feature:
\begin{equation}
\big[\mathrm{Bilinear}(\mathbf{a}, \mathbf{b})\big]_k = \mathbf{a}^\top \mathbf{W}_k\, \mathbf{b} + \beta_k, \quad k = 1,\dots,d',
\label{eq:bilinear}
\end{equation}
where $\mathbf{W}_k\in\mathbb{R}^{d\times d}$ contains learned weights and $\beta_k$ is a bias. Each entry of $\mathbf{W}_k$ weights a product of one component of $\mathbf{a}$ and one of $\mathbf{b}$. Their weighted sum plus the bias gives one component of the pair feature. The $d'=d/2$ weight matrices capture different combinations of the endpoint features. A ReLU and a linear layer reduce this feature to a score, and a sigmoid gives the connection probability.

\subsection{Four-Stage Connectivity Prediction}\label{ssec:cascade}

\begin{figure*}[!t]
\centering
\includegraphics[width=\textwidth]{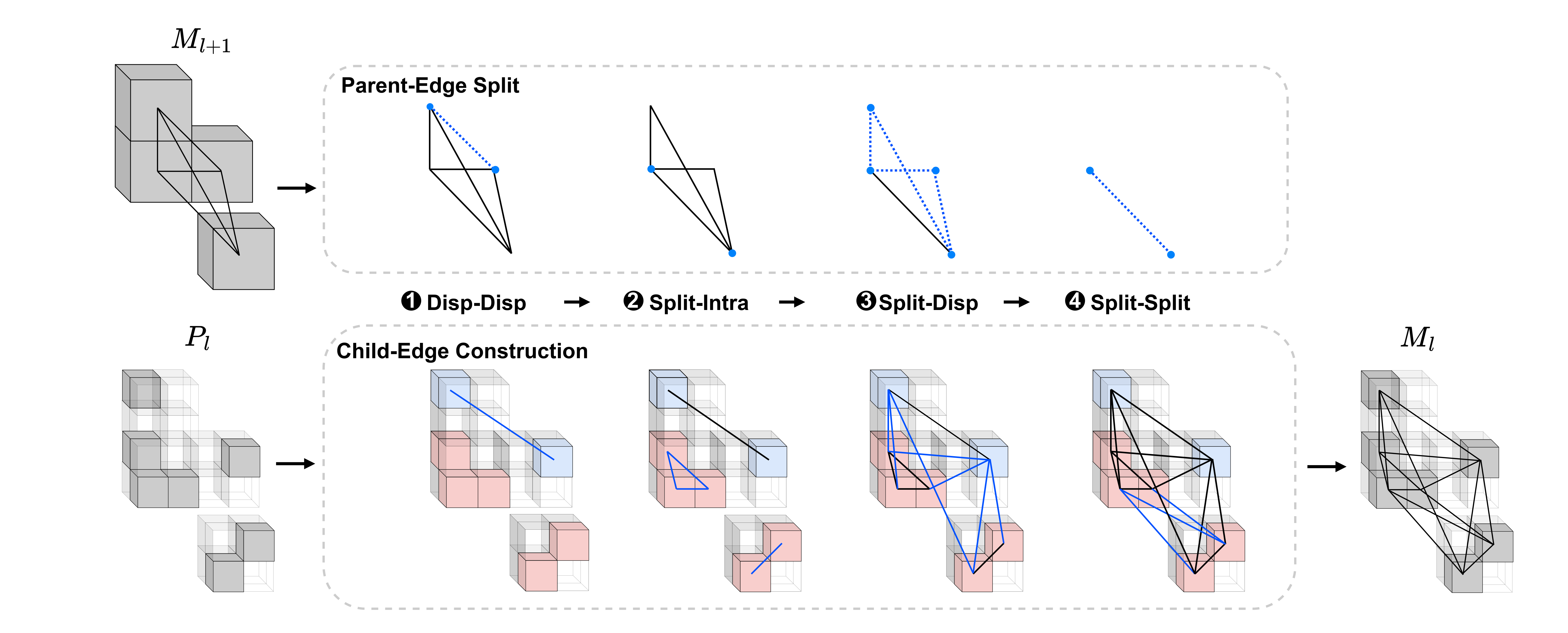}
\caption{Child connectivity is built progressively on the decoded vertices $P_l$. Disp-Disp first inherits connections from the parent mesh $M_{l+1}$. Split-Intra adds connections within each Split parent, after which Split-Disp and Split-Split determine how the remaining parent edges expand into child connections. Blue dotted lines highlight the parent edges being processed, while blue solid lines show the newly added child edges. These child edges are retained in black in subsequent stages, gradually forming the connectivity of $M_l$. An interactive version of this process is available online (see footnote~\ref{fn:demo}).}
\label{fig:pipeline}
\end{figure*}

Figure~\ref{fig:pipeline} summarizes the four-stage prediction process. Disp-Disp first recovers inherited connections directly from the parent mesh. Split-Intra then predicts connections within each Split parent. Its binarized output provides internal connectivity context for the subsequent Split-Disp and Split-Split stages, which predict connections between different parents. Following this order, the stages progressively add child edges to the decoded vertex set $P_l$.

As a complementary view, the inter-parent stages predict \emph{how parent edges split into child edges}. A parent edge's splitting pattern specifies which children at its two ends are connected. Disp-Disp inherits a single connection, while Split-Disp and Split-Split predict the pattern when one or both endpoints expand into multiple children. Split-Intra supplies the connections inside these expanded parents, providing local structure for predicting the inter-parent patterns.

\subsection{Parent Feature Extraction: GraphAware-PFE}\label{ssec:pfe}
To build the parent features $\mathbf{g}^{(p')}$ used by the three predictors, GraphAware-PFE combines three attention paths (Fig.~\ref{fig:netarch}a). KNN attention gathers information from spatial neighbors, graph attention from vertices connected by parent edges, and global attention from the parent mesh as a whole.

At each level, it takes the coordinates of the $N_{l+1}=|P_{l+1}|$ parent vertices and the edges $E_{l+1}$ as input. A two-layer MLP maps the coordinates to $d$-dimensional features. The $K=12$ nearest neighbors of each vertex are computed once and shared by four GraphAwareBlocks, each containing the three attention paths.

\emph{KNN attention} aggregates features from nearby vertices. Relative coordinates are included in the attention keys and values, following the use of relative position information in Point Transformer~\cite{zhao2021pt}:
\begin{equation}
\mathbf{a}^{\text{local}}_i = \sum_{j \in \mathrm{knn}(i)} \alpha_{ij}\, \mathbf{W}_v\,[\mathbf{h}_j \,\|\, \mathbf{r}_{ij}],
\end{equation}
where $\mathrm{knn}(i)$ is the set of $K$ nearest neighbors of vertex $i$ and $\alpha_{ij}$ is the softmax attention weight. The vector $\mathbf{h}_j$ is the feature of neighbor $j$, $\mathbf{r}_{ij}$ is its coordinate relative to $i$, and $\mathbf{W}_v$ is a learned projection of their concatenation.

\emph{Graph attention} aggregates along the parent edges $E_{l+1}$ with softmax weights~\cite{velickovic2018gat}. It complements KNN attention because spatial proximity and mesh adjacency define different neighborhoods. Its aggregation cost is $O(|E_{l+1}|)$.

\emph{Global attention} exchanges information through $G=4$ learnable tokens, inspired by Set Transformer~\cite{lee2019set} and Perceiver~\cite{jaegle2021perceiver}. These tokens are shared feature vectors that collect and redistribute information across the parent mesh. Vertices attend to the tokens, and the tokens attend to all vertices to update their state. This exchange costs $O(N_{l+1}G)$ per block.

The three path outputs are added to the block input and refined by a feed-forward network. The global tokens are also updated residually across blocks. After the last block, LayerNorm and a linear projection produce $\mathbf{g}^{(p')}$. For fixed feature width and precomputed neighborhoods, attention aggregation costs $O\!\big(N_{l+1}(K+G)+|E_{l+1}|\big)$ per block, avoiding the all-pairs interactions of full vertex self-attention.

\subsection{Edge Coverage and Level-Aware Prediction}\label{ssec:level-aware}

\begin{figure}[!t]
\centering
\includegraphics[width=0.85\columnwidth]{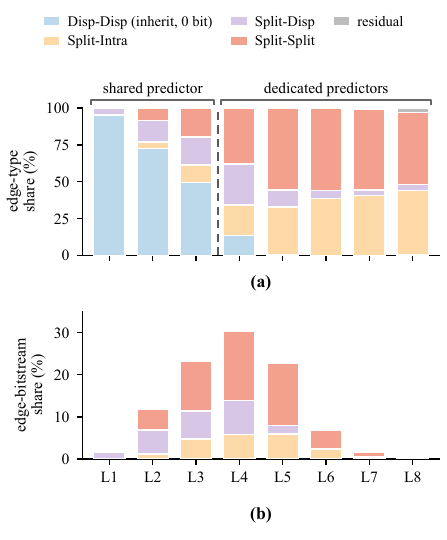}
\caption{Per-level statistics of the connectivity branch. (a) Edge counts as percentages of all edges at each level, including residuals in gray; each bar sums to approximately $100\%$. The dashed line separates the shared branch for $L_1$--$L_3$ from the dedicated branches for $L_4$--$L_8$. (b) Each bar gives that level's share of the refinement connectivity bits, stacked by category; all bars together sum to $100\%$. Disp-Disp contributes no bits because its edges are inherited.}
\label{fig:perlevel}
\end{figure}

Edge distributions change with scale. In Fig.~\ref{fig:perlevel}a, Disp-Disp accounts for about $95\%$ of the edges at $L_1$, while Split-Split becomes dominant at coarser levels. Figure~\ref{fig:perlevel}b shows that $L_3$--$L_5$ account for about $76.6\%$ of the refinement connectivity bits.

A retained parent edge generates every pair formed by one child at each endpoint. All true child edges between these two parent groups are therefore covered. Several fine-level triangles can produce the same parent edge during pooling. If one triangle collapses, another triangle can still preserve that edge, so the associated child candidates remain available.

The gray residual portion in Fig.~\ref{fig:perlevel}a contains the exceptions: some parent connections disappear because all triangles that could preserve them collapse. True child edges between those parent groups then fall outside the candidates. Pooled across the 256 evaluation frames and their refinement levels, candidate coverage is $99.9795\%$, or $99.9548\%$ after excluding inherited Disp-Disp edges. Uncovered edges are transmitted as endpoint-index pairs and merged with the inherited and arithmetic-decoded edges. This completes edge recovery at each level, even where candidate coverage is lower.

These scale-dependent distributions motivate level-aware probability models. In our nine-level hierarchy, the transitions reconstructing $L_4$--$L_8$ use five independent connectivity branches, while those reconstructing $L_1$--$L_3$ share one branch. The dashed line in Fig.~\ref{fig:perlevel} marks this boundary. Each branch contains GraphAware-PFE and the three neural predictors with the same architecture; only the learned parameters differ. Disp-Disp retains the same inheritance rule.

\section{Experiments}\label{sec:exp}
\providecommand{\OMExpTableStyle}{%
  \small
  \setlength{\tabcolsep}{3pt}%
  \renewcommand{\arraystretch}{1.15}%
  \setlength{\heavyrulewidth}{0.8pt}%
  \setlength{\lightrulewidth}{0.5pt}}

\providecommand{\OMDetailTableStyle}{%
  \small
  \setlength{\tabcolsep}{4pt}%
  \renewcommand{\arraystretch}{1.25}%
  \setlength{\heavyrulewidth}{1pt}%
  \setlength{\lightrulewidth}{0.45pt}%
  \setlength{\aboverulesep}{0.5ex}%
  \setlength{\belowrulesep}{0.75ex}}
\providecommand{\OMClassicHead}[2][c]{\makecell[#1]{\bfseries #2}}
\providecommand{\OMClassicNum}[1]{\makebox[2.8em][r]{#1}}
\providecommand{\OMDraftHead}[1]{\shortstack[c]{#1}}
\providecommand{\OMDraftNote}[1]{%
  \par\vspace{4pt}\begin{minipage}{\columnwidth}\footnotesize\raggedright #1\end{minipage}}

\begin{table*}[!t]
\centering
\caption{Bitrate Comparison on Eight Sequences (BPF)}
\label{tab:main}
\normalsize
\renewcommand{\arraystretch}{1.16}
\setlength{\tabcolsep}{4pt}
\begin{tabular*}{0.95\textwidth}{@{\extracolsep{\fill}}lccccc@{}}
\toprule
Sequence & \makecell{meshopt.\\+zstd~\cite{meshopt}} & \makecell{SC3DMC\\\cite{mamou2009}} & \makecell{Draco\\\cite{draco}} & \makecell{V-Mesh\\\cite{iso23090}} & \makecell{\textbf{OctMesh}\\\textbf{(Ours)}} \\
\midrule
basketball\_player & 16.059 & 10.300 & 9.128 & 8.679 & \textbf{7.681} \\
dancer & 15.826 & 10.013 & 8.990 & 8.451 & \textbf{7.478} \\
football & 15.936 & 9.713 & 9.039 & 8.405 & \textbf{7.539} \\
levi & 15.858 & 9.382 & 8.360 & 7.942 & \textbf{6.881} \\
longdress & 13.399 & 7.270 & 5.790 & 6.099 & \textbf{5.281} \\
mitch & 16.842 & 10.804 & 9.505 & 9.371 & \textbf{8.642} \\
soldier & 13.844 & 7.459 & 5.701 & 6.419 & \textbf{5.506} \\
thomas & 16.651 & 10.080 & 8.848 & 9.163 & \textbf{7.257} \\
\midrule
Avg. geometry & 9.069 & 7.480 & 6.993 & 6.765 & \textbf{4.650} \\
Avg. connectivity & 6.478 & 1.884 & 1.174 & 1.240 & 2.167 \\
Avg. overhead & 0.004 & 0.013 & 0.003 & 0.061 & 0.216 \\
\midrule
\textbf{Avg total (BPF)} $\downarrow$ & 15.552 & 9.378 & 8.170 & 8.066 & \textbf{7.033} \\
\bottomrule
\end{tabular*}
\end{table*}

\subsection{Experimental Setup}\label{ssec:setup}\label{sec:impl}

We train OctMesh on the UVG-VM dataset~\cite{uvgvm} and the exercise and model sequences of Owlii~\cite{owlii}. For evaluation, we use eight voxelized human sequences from the MPEG V-DMC test material~\cite{iso23090}: basketball\_player, dancer, football, levi, longdress, mitch, soldier and thomas. Each evaluation frame has 10-bit or 12-bit coordinates. We voxelize each mesh, merge vertices that share the same voxel coordinates, and remove isolated vertices; the resulting mesh serves as the common reference $M_1$ for all lossless comparisons, while texture coordinates and normals are outside our coding scope. From $M_1$, OctMesh builds $M_2,\ldots,M_9$ by the octree pooling and connectivity update of Sec.~\ref{ssec:pooling}, and every baseline receives the same $M_1$.

Unless otherwise noted, experiments are conducted on a server with an AMD EPYC 7543 CPU and two 32GB NVIDIA GeForce RTX 5090 D GPUs. We use PyTorch, spconv~\cite{yan2018}, torchac~\cite{mentzer2019}, pytorch3d~\cite{pytorch3d} and torch-geometric~\cite{pyg}. We retain the geometry architecture of Unicorn~\cite{wang2024}, set its feature width to 32 and retrain it on the training data above. The connectivity branch uses width $d=64$. Each Octant Transformer has two layers and four heads. GraphAware-PFE has four blocks with four heads each, $K=12$ spatial neighbors and $G=4$ global tokens. The level-specific and shared branches follow Sec.~\ref{ssec:level-aware}. The model has $7.54$M parameters.

We first train the geometry branch, then freeze it and train the connectivity branch. The geometry branch uses Adam with a learning rate of $5\times10^{-5}$. The connectivity branch uses AdamW with a weight decay of $10^{-5}$; during continued training, its learning rate decays from $5\times10^{-5}$ to $2.5\times10^{-5}$. Both branches use a batch size of 4 and gradient clipping with a maximum norm of 1. The geometry and connectivity branches are trained for $4.25\times10^5$ and $4.5\times10^5$ steps, respectively.

Both branches are trained using binary cross-entropy (BCE) losses. The geometry branch predicts whether an octant is occupied, and the connectivity branch predicts whether a candidate edge exists. Let $y_i\in\{0,1\}$ be a target symbol and $\hat{y}_i$ the predicted probability that it equals 1. The BCE for this symbol is
\begin{equation}
\BCE(y_i,\hat{y}_i)=-y_i\ln\hat{y}_i-(1-y_i)\ln(1-\hat{y}_i).
\label{eq:bce}
\end{equation}
The geometry loss sums BCE over occupancy symbols across the training hierarchy and divides this sum by the number of vertices at its finest level. For connectivity, we first average BCE over the valid candidate edges within each prediction stage at each level. Denote this mean by $\ell_{l,s}$, where $l$ indexes the refinement level and $s\in\{\mathrm{SI},\mathrm{SD},\mathrm{SS}\}$ denotes Split-Intra, Split-Disp or Split-Split, respectively. We then add the three stage means at each level and average the resulting sums over levels with at least one valid candidate in these stages. Writing this set of levels as $\mathcal{T}$, the connectivity loss is
\begin{equation}
\mathcal{L}_{\mathrm{edge}}=\frac{1}{|\mathcal{T}|}\sum_{l\in\mathcal{T}}
\bigl(\ell_{l,\mathrm{SI}}+\ell_{l,\mathrm{SD}}+\ell_{l,\mathrm{SS}}\bigr).
\label{eq:branch-loss}
\end{equation}
A stage with no valid candidates contributes zero, and the three stage means have equal weight within each level. Disp-Disp and residual edges are excluded from this loss because neither uses a learned probability model.

We verify OctMesh's reconstruction by comparing the decoded vertex-coordinate, edge, and face sets with the reference. We report bits per face, $\mathrm{BPF}=B/|F_1|$, where $B$ counts all transmitted bits and $|F_1|$ is the face count of $M_1$. In Table~\ref{tab:main}, we first average frame-wise BPF over the same 32 frames per sequence for OctMesh and all baselines, then average these sequence means to obtain the overall rates.

We compare OctMesh with four codecs. The V-Mesh reference implementation~\cite{iso23090} uses its all-intra configuration, with position quantization matched to the input bit depth: 12 for six sequences and 10 for longdress and soldier. MPEG-4 SC3DMC, implemented by Open3DGC~\cite{open3dgc}, uses TFAN connectivity coding~\cite{mamou2009} and parallelogram geometry prediction with 13-bit quantization. Draco 1.5.7~\cite{draco} uses compression level 10 and position quantization of 12 or 10 bits for the same sequence groups. Coordinate checks for Draco and SC3DMC use integer-grid rounding. Meshoptimizer v0.25~\cite{meshopt} encodes 16-bit integer vertex coordinates and the triangle index buffer and compresses the two payloads separately with zstd at level 19; a 20-byte container header is included. DMGC~\cite{zhao2023} is not included in the experimental comparison because no public implementation was available at the time of evaluation.

\subsection{Lossless Compression Results}
Table~\ref{tab:main} reports the rates on eight sequences. OctMesh recovers the finest-level mesh at $7.033$ BPF while retaining the nine-level vertex-and-edge hierarchy, approximately $12.8\%$ below V-Mesh, the strongest total-rate baseline in this comparison.

\textbf{Geometry and connectivity rates.} The gain comes mainly from geometry: OctMesh uses $4.650$ BPF, versus $6.765$ BPF for V-Mesh. Coarse vertices constrain the local child-occupancy patterns to be coded. OctMesh's $2.167$ connectivity BPF includes $2.004$ for hierarchical refinements and $0.163$ for base-layer faces. Its connectivity cost exceeds the single-level baselines, but the $2.116$ BPF geometry saving over V-Mesh offsets the $0.927$ BPF connectivity increase.

\textbf{Overhead and face recovery.} Overhead comprises stream headers ($0.161$ BPF) and a compressed selection mask that identifies the finest-level unoriented faces among canonically ordered triangular cycles of the decoded edge graph. The face payload, including its header, adds $0.055$ BPF, yielding $0.216$ BPF overhead and $7.033$ BPF in total. The decoder needs no reference face list.

\subsection{Progressive Reconstruction and Lossy Comparison}\label{ssec:lod-metric}

The vertex-and-edge stream supports progressive refinement. The intermediate surfaces use faces from the reference hierarchy, and their rates count cumulative geometry-and-edge records. Face-selection information is included only at the finest level.

\begin{figure}[!t]
\centering
\includegraphics[width=\columnwidth]{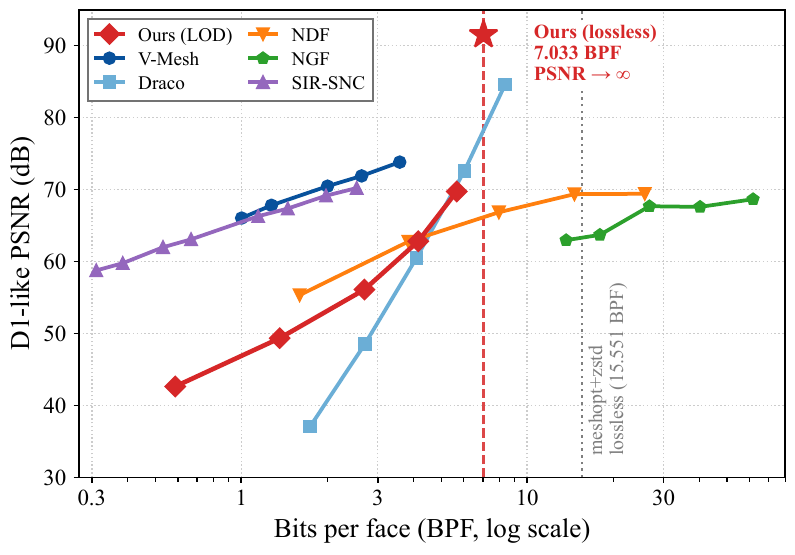}
\caption{Rate--distortion curves using D1-like PSNR, averaged over representative frames. The red line shows hierarchy previews at cumulative geometry-and-edge rates. The red dashed line marks its average lossless rate in Table~\ref{tab:main} ($7.033$ BPF), and the star illustrates the ideal zero-distortion limit ($\mathrm{PSNR}\to\infty$). The gray dotted line marks meshoptimizer+zstd's lossless rate.}
\label{fig:rd-lod}
\end{figure}

\textbf{Comparison protocol.} We compare OctMesh's hierarchy previews with lossy V-Mesh and Draco, and with NDF~\cite{ndf2025}, NGF~\cite{ngf2024} and SIR-SNC~\cite{sirsnc2026}. NDF and NGF use five rate settings, while SIR-SNC uses pretrained models at four resolutions. 
Following the use of point-to-surface distances in learned mesh compression~\cite{chen2023npm,ndf2025}, we assess the geometric fidelity of progressive reconstructions and lossy baselines using a bidirectional point-to-surface measure. Each reference mesh is translated and scaled so that its longest bounding-box edge is one; the reconstruction receives the same transform. We take the larger directional mean as $D$ and report $-20\log_{10}D$ in decibels. OctMesh, V-Mesh and Draco use 48 nearest-centroid candidate triangles with original $L_1$ reference meshes; NDF, NGF and SIR-SNC use a PyTorch3D-based evaluator with manifold-preprocessed references.

\textbf{Progressive rate--distortion behavior.} OctMesh provides a vertex-and-edge reconstruction after each decoded refinement level. The displayed LODs span approximately $0.6$--$5.7$ BPF; for example, $1.4$ BPF gives roughly 6k vertices. The complete representation also recovers the finest-level unoriented faces. Figure~\ref{fig:rd-lod} compares the intermediate reconstructions with the lossy operating points of the baselines.

V-Mesh reaches about 66 dB at $1$ BPF, and SIR-SNC reaches 70 dB below $3$ BPF. V-Mesh reaches about 74 dB at $3.6$ BPF. NDF's two highest-rate points are both near 69.4 dB, while Draco improves as its quantization becomes finer. The OctMesh curve shows how the hierarchy refines geometry across levels; the lossy curves provide reference operating points under the evaluation protocols described above.

\begin{figure*}[!t]
\centering
\includegraphics[width=0.85\textwidth]{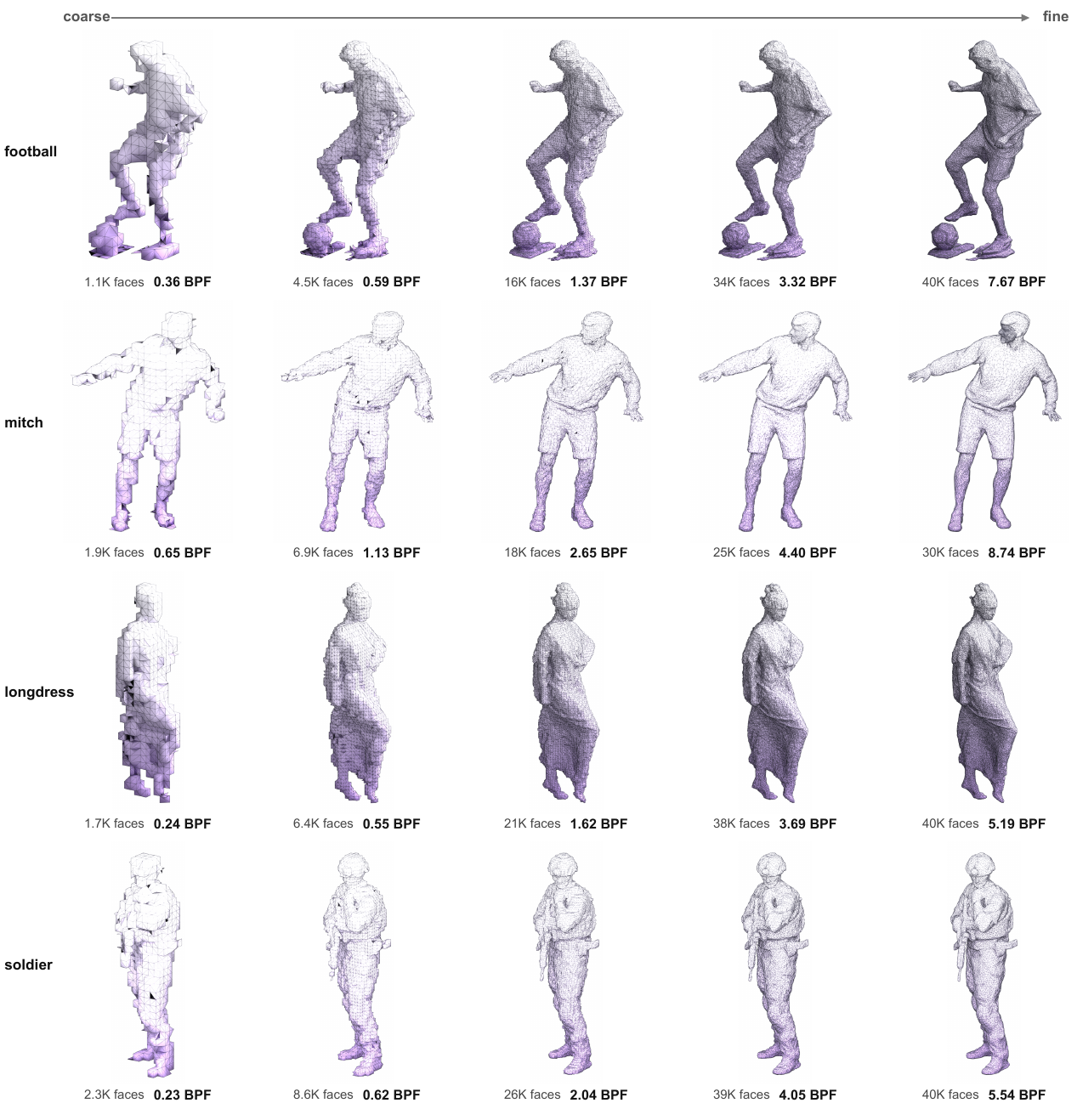}
\caption{Reference hierarchy views for football, mitch, longdress and soldier, from coarse to fine. Labels give face counts and cumulative geometry-and-edge BPF; the rightmost $L_1$ rates additionally include face selection.}
\label{fig:lod}
\end{figure*}

\textbf{Qualitative progression.} In the reference hierarchy views of Fig.~\ref{fig:lod}, coarse levels preserve the overall pose, while successive refinements recover finer surface details before reaching $L_1$. The labels show the corresponding cumulative rates.

\subsection{Coding Cost and Runtime}
We examine the coding cost of each stage and the computational cost of the model.

\textbf{Stage-wise coding performance.} Table~\ref{tab:abl-stage-cost} summarizes the 256 evaluation frames. GT share is the fraction of true edges among candidates; precision is the fraction of true edges among positive predictions. From Split-Intra to Split-Split, candidate counts increase while GT share and precision decrease. Split-Split accounts for about $53\%$ of refinement bits, making it the main coding cost. Disp-Disp inherits edges without bits. Residual edges whose parent connections disappear during pooling are transmitted as child-index pairs and restored at decoding, costing $0.033$ BPF. Arithmetic decoding recovers actual symbols regardless of prediction errors.

\begin{table}[!t]
\centering
\caption{Per-Stage Prediction Quality and Bitrate of the Connectivity Branch}
\label{tab:abl-stage-cost}
\OMExpTableStyle
\begin{tabular*}{\columnwidth}{@{\extracolsep{\fill}}lrrrrr@{}}
\toprule
Stage & Cand. & GT share & Precision & Edge BPF & Bit share \\
\midrule
Split-Intra & 8.8\,M  & 74.1\% & 89.3\% & 0.408 & 20\% \\
Split-Disp  & 15.4\,M & 53.2\% & 86.7\% & 0.494 & 25\% \\
Split-Split & 59.4\,M & 18.7\% & 72.9\% & 1.069 & 53\% \\
Residual & --- & --- & --- & 0.033 & 2\% \\
\addlinespace[3pt]
\textbf{Total} & & & & \textbf{2.004} & 100\% \\
\bottomrule
\end{tabular*}
\end{table}

\begin{table}[!t]
\centering
\caption{Model Complexity and Runtime Breakdown}
\label{tab:efficiency}
\OMDetailTableStyle
\textnormal{(a) Model complexity}\par\vspace{3pt}
\begin{tabular}{>{\raggedright\arraybackslash}p{\dimexpr 0.54\columnwidth-8pt\relax}>{\centering\arraybackslash}p{\dimexpr 0.23\columnwidth-8pt\relax}>{\centering\arraybackslash}p{\dimexpr 0.23\columnwidth-8pt\relax}}
\toprule
\OMClassicHead[l]{Module} & \OMClassicHead{Params\\(M)} & \OMClassicHead{MFLOPs\\ /face} \\
\midrule
GraphAware-PFE & \OMClassicNum{0.41} & \OMClassicNum{2.18} \\
Split-Intra & \OMClassicNum{0.17} & \OMClassicNum{1.87} \\
Split-Disp & \OMClassicNum{0.17} & \OMClassicNum{1.70} \\
Split-Split & \OMClassicNum{0.46} & \OMClassicNum{2.77} \\
\midrule[0.35pt]
Connectivity branch & \OMClassicNum{7.26} & \OMClassicNum{8.52} \\
\textbf{Total} & \OMClassicNum{\textbf{7.54}} & \OMClassicNum{\textbf{13.40}} \\
\bottomrule
\end{tabular}
\par\vspace{9pt}
\textnormal{(b) Encoding and decoding time}\par\vspace{3pt}
\begin{tabular}{>{\raggedright\arraybackslash}p{\dimexpr 0.54\columnwidth-8pt\relax}>{\centering\arraybackslash}p{\dimexpr 0.23\columnwidth-8pt\relax}>{\centering\arraybackslash}p{\dimexpr 0.23\columnwidth-8pt\relax}}
\toprule
\OMClassicHead[l]{Module} & \OMClassicHead{Encoding\\(s/frame)} & \OMClassicHead{Decoding\\(s/frame)} \\
\midrule
GraphAware-PFE & \OMClassicNum{0.2120} & \OMClassicNum{0.2105} \\
Split-Intra & \OMClassicNum{0.0657} & \OMClassicNum{0.0537} \\
Split-Disp & \OMClassicNum{0.2011} & \OMClassicNum{0.1404} \\
Split-Split & \OMClassicNum{0.3433} & \OMClassicNum{0.3084} \\
\midrule[0.35pt]
Connectivity subtotal & \OMClassicNum{0.8220} & \OMClassicNum{0.7131} \\
Geometry branch & \OMClassicNum{0.8161} & \OMClassicNum{0.8256} \\
Others & \OMClassicNum{0.3372} & \OMClassicNum{0.0408} \\
\midrule[0.35pt]
\textbf{Total} & \OMClassicNum{\textbf{1.975}\phantom{0}} & \OMClassicNum{\textbf{1.579}\phantom{0}} \\
\bottomrule
\end{tabular}
\end{table}

\textbf{Model size and computational cost.} In Table~\ref{tab:efficiency}(a), the first four parameter counts describe one level group; the connectivity subtotal covers all six groups. The codec has $7.54$M parameters and costs approximately $488$ GFLOPs per frame ($13.40$ MFLOPs/face: $8.52$ for connectivity and $4.88$ for geometry). One multiply--accumulate counts as two FLOPs; custom sparse-convolution and KNN kernels are excluded.

\textbf{Measured encoding and decoding time.} Table~\ref{tab:efficiency}(b) reports component timings and total runtime on the same server. The connectivity subtotal covers the four listed neural modules; Geometry branch includes arithmetic coding, and Others covers layer ordering, residuals, and base-layer handling. The 288-frame averages are $1.975$ s/frame for encoding and $1.579$ s/frame for decoding; bitrate statistics use 256 frames. Component times use four decimal places and totals use three. Face-selection time is excluded.

\subsection{Connectivity Modeling Ablations}
\label{ssec:controlled-connectivity}
\label{ssec:connectivity-ablations}

Table~\ref{tab:connectivity-ablation} combines two comparisons on the 256-frame evaluation set. The full OctMesh model, with all attention paths and specialized entropy heads, gives $2.167$ connectivity BPF in both comparisons at the reported precision.

\begin{table}[!t]
\centering
\caption{Connectivity Modeling Ablations}
\label{tab:connectivity-ablation}
\label{tab:abl-pfe}
\label{tab:controlled-results}
\label{tab:controlled-structure}
\OMDetailTableStyle
\begin{tabular}{>{\raggedright\arraybackslash}p{\dimexpr 0.66\columnwidth-8pt\relax}>{\centering\arraybackslash}p{\dimexpr 0.34\columnwidth-8pt\relax}}
\toprule
\OMClassicHead[l]{Configuration} & \OMClassicHead{Conn. BPF $\downarrow$} \\
\midrule
\multicolumn{2}{l}{\textit{(a) Attention paths (fixed weights)}} \\
KNN only & \OMClassicNum{4.370} \\
KNN + graph & \OMClassicNum{4.397} \\
KNN + global & \OMClassicNum{2.268} \\
\addlinespace[4pt]
\multicolumn{2}{l}{\textit{(b) Entropy models (shared geometry)}} \\
Level-and-type prior & \OMClassicNum{6.832} \\
Shared pairwise MLP & \OMClassicNum{4.893} \\
\midrule[0.35pt]
\textbf{OctMesh (full model)} & \OMClassicNum{\textbf{2.167}} \\
\bottomrule
\end{tabular}
\end{table}

\textbf{Attention paths.} We fix the trained weights and disable selected GraphAware-PFE outputs. These archived mask evaluations use an NVIDIA GeForce RTX 3090. KNN + global gives $2.268$ BPF, $48.1\%$ below KNN only ($4.370$ BPF), whereas KNN + graph is slightly higher at $4.397$ BPF. Retaining all three paths reduces the rate by $50.4\%$ relative to KNN only and by a further $4.5\%$ relative to KNN + global. The trained model is therefore more sensitive to removing global attention than graph attention. Graph attention still improves the full configuration, but its benefit is conditional on the other retained paths. These inference-time masks measure dependence of the trained model; the reduced architectures are not retrained.

\textbf{Entropy models.} The level-and-type prior, shared pairwise MLP, and full OctMesh use the same geometry bitstream, hierarchy, candidates, Disp-Disp inheritance, residual representation, entropy coder, and reconstruction target. Frame selection, bit accounting, and averaging follow Sec.~\ref{ssec:setup}; the full-model operating point is that of Table~\ref{tab:main}. The non-neural prior uses $p_{l,t}=(N^1_{l,t}+1)/(N^1_{l,t}+N^0_{l,t}+2)$, where the counts are present and absent training candidates at level $l$ and category $t\in\{\mathrm{SI},\mathrm{SD},\mathrm{SS}\}$. Its probabilities are frozen before evaluation, without instance-dependent or predicted SI context. The shared MLP retains GraphAware-PFE and the level-parameter groups, replacing the three specialized heads with one $171\!-\!256\!-\!256\!-\!128\!-\!1$ MLP per group (ReLU hidden layers; sigmoid output). It takes endpoint-parent features, occupancy masks, local octant coordinates, edge category, and predicted SI context with availability flags. SI prediction receives zero context; SD and SS use its own SI predictions thresholded at $p\geq0.5$, with unavailable context zeroed and flagged.

The MLP is trained once with geometry frozen, using the loss, batch size, optimizer, and gradient clipping of Sec.~\ref{ssec:setup}. Its learning rate decays linearly from $5\times10^{-5}$ to $2.5\times10^{-5}$ over 450k updates. Validation runs every 10k updates; the checkpoint with the lowest actual connectivity BPF is selected, with ties favoring the earlier one. The shared MLP saves $28.4\%$ relative to the prior; OctMesh further saves $55.7\%$ relative to the MLP and $68.3\%$ relative to the prior. Because the neural models share the PFE architecture, level grouping, and access to predicted SI context, this comparison assesses the specialized prediction modules collectively. The common candidate partition is retained, so its contribution is not isolated.

\FloatBarrier
\section{Conclusion}
This paper proposes OctMesh, a learned framework for lossless triangle mesh compression on a multiscale octree hierarchy. Our core contributions are:
(i) observing that octree downsampling naturally divides parent vertices into Displacement and Split types, and thereby partitioning child-level edges exactly into four disjoint categories;
(ii) designing the corresponding four-stage predictors, from zero-bit inference (Disp-Disp) to $8\times 8$ neural prediction (Split-Split), maximizing prior reuse through a cascaded context;
(iii) proposing GraphAware-PFE to extract high-quality parent-vertex features.
On the eight MPEG V-DMC sequences, OctMesh recovers the finest-level vertex, edge and unoriented face sets at 7.033 BPF while retaining nine levels of vertex-and-edge refinement, approximately $12.8\%$ below V-Mesh. By organizing geometry and edge coding on the same octree hierarchy, OctMesh brings multiscale neural compression to meshes and supports progressive geometric refinement.



\begin{thebibliography}{99}
\bibitem{rossignac1999}
J. Rossignac, ``Edgebreaker: Connectivity Compression for Triangle Meshes,'' \emph{IEEE Trans. Vis. Comput. Graph.}, 1999.

\bibitem{touma1998}
C. Touma and C. Gotsman, ``Triangle Mesh Compression,'' in \emph{Graphics Interface}, 1998.

\bibitem{alliez2001}
P. Alliez and M. Desbrun, ``Valence-Driven Connectivity Encoding for 3D Meshes,'' in \emph{Eurographics}, 2001.

\bibitem{draco}
Google, \emph{Draco 3D Graphics Compression Library}, \url{https://github.com/google/draco}.

\bibitem{iso23090}
ISO/IEC 23090-29, \emph{Information Technology---Coded Representation of Immersive Media---Part 29: Video-Based Dynamic Mesh Coding (V-DMC)}, ISO/IEC JTC 1/SC 29, 2026.

\bibitem{marais2007distance}
P. Marais, J. Gain, and D. Shreiner, ``Distance-Ranked Connectivity Compression of Triangle Meshes,'' \emph{Computer Graphics Forum}, vol. 26, no. 4, pp. 813--823, 2007.

\bibitem{dvorak2023priority}
J. Dvo\v{r}\'{a}k, Z. K\'{a}\v{c}erekov\'{a}, P. Van\v{e}\v{c}ek, and L. V\'{a}\v{s}a, ``Priority-Based Encoding of Triangle Mesh Connectivity for a Known Geometry,'' \emph{Computer Graphics Forum}, vol. 42, no. 1, pp. 60--71, 2023.

\bibitem{hoppe1996}
H. Hoppe, ``Progressive Meshes,'' in \emph{ACM SIGGRAPH}, 1996.

\bibitem{karni2000}
Z. Karni and C. Gotsman, ``Spectral Compression of Mesh Geometry,'' in \emph{ACM SIGGRAPH}, 2000.

\bibitem{khodakovsky2000}
A. Khodakovsky, P. Schr\"{o}der, and W. Sweldens, ``Progressive Geometry Compression,'' in \emph{ACM SIGGRAPH}, 2000.

\bibitem{zhao2023}
X. Zhao, X. Zeng, L. Gao, Y. Xu, and Y. Wang, ``DMGC: Deep Triangle Mesh Geometry Compression via Connectivity Prediction,'' in \emph{ACM NOSSDAV}, 2023.

\bibitem{huang2020}
L. Huang et al., ``OctSqueeze: Octree-Structured Entropy Model for LiDAR Compression,'' in \emph{CVPR}, 2020.

\bibitem{fu2022}
C. Fu et al., ``OctAttention: Octree-Based Large-Scale Contexts Model for Point Cloud Compression,'' in \emph{AAAI}, 2022.

\bibitem{wang2024}
J. Wang et al., ``A Versatile Point Cloud Compressor Using Universal Multiscale Conditional Coding -- Part I: Geometry,'' \emph{IEEE TPAMI}, 2025.

\bibitem{wang2023}
J. Wang et al., ``Sparse Tensor-Based Multiscale Representation for Point Cloud Geometry Compression,'' \emph{IEEE TPAMI}, 2023.

\bibitem{yan2018}
Y. Yan et al., ``SECOND: Sparsely Embedded Convolutional Detection,'' \emph{Sensors}, 2018.

\bibitem{choy2019}
C. Choy et al., ``4D Spatio-Temporal ConvNets: Minkowski Convolutional Neural Networks,'' in \emph{CVPR}, 2019.

\bibitem{sirsnc2026}
J. Wang, S. Ren, and J. Hou, ``Neural Compression of 3D Meshes Using Sparse Implicit Representation,'' in \emph{ICLR}, 2026.

\bibitem{peng2005survey}
J. Peng, C.-S. Kim, and C.-C. J. Kuo, ``Technologies for 3D Mesh Compression: A Survey,'' \emph{J. Vis. Commun. Image Represent.}, 2005.

\bibitem{maglo2015}
A. Maglo, G. Lavou\'{e}, F. Dupont, and C. Hudelot, ``3D Mesh Compression: Survey, Comparisons, and Emerging Trends,'' \emph{ACM Computing Surveys}, 2015.

\bibitem{deering1995}
M. Deering, ``Geometry Compression,'' in \emph{ACM SIGGRAPH}, 1995.

\bibitem{taubin1998}
G. Taubin and J. Rossignac, ``Geometric Compression Through Topological Surgery,'' \emph{ACM Trans. Graphics}, 1998.

\bibitem{gumhold1998}
S. Gumhold and W. Stra{\ss}er, ``Real Time Compression of Triangle Mesh Connectivity,'' in \emph{ACM SIGGRAPH}, 1998.

\bibitem{isenburg2000facefixer}
M. Isenburg and J. Snoeyink, ``Face Fixer: Compressing Polygon Meshes with Properties,'' in \emph{ACM SIGGRAPH}, 2000.

\bibitem{mamou2009}
K. Mamou, T. Zaharia, and F. Pr\^{e}teux, ``TFAN: A Low Complexity 3D Mesh Compression Algorithm,'' \emph{Computer Animation and Virtual Worlds}, 2009.

\bibitem{mammou2022vdmc}
K. Mammou, J. Kim, A. M. Tourapis, D. Podborski, and D. Flynn, ``Video and Subdivision Based Mesh Coding,'' in \emph{EUVIP}, 2022.

\bibitem{pajarola2000}
R. Pajarola and J. Rossignac, ``Compressed Progressive Meshes,'' \emph{IEEE Trans. Vis. Comput. Graph.}, 2000.

\bibitem{alliez2001progressive}
P. Alliez and M. Desbrun, ``Progressive Compression for Lossless Transmission of Triangle Meshes,'' in \emph{ACM SIGGRAPH}, 2001.

\bibitem{valette2009ipr}
S. Valette, R. Chaine, and R. Prost, ``Progressive Lossless Mesh Compression via Incremental Parametric Refinement,'' \emph{Computer Graphics Forum}, 2009.

\bibitem{balle2018}
J. Ball\'{e}, D. Minnen, S. Singh, S. J. Hwang, and N. Johnston, ``Variational Image Compression with a Scale Hyperprior,'' in \emph{ICLR}, 2018.

\bibitem{minnen2018}
D. Minnen, J. Ball\'{e}, and G. Toderici, ``Joint Autoregressive and Hierarchical Priors for Learned Image Compression,'' in \emph{NeurIPS}, 2018.

\bibitem{he2021checkerboard}
D. He, Y. Zheng, B. Sun, Y. Wang, and H. Qin, ``Checkerboard Context Model for Efficient Learned Image Compression,'' in \emph{CVPR}, 2021.

\bibitem{zhu2024dynamic}
L. Zhu, B. Li, R. Lu, P. Chen, Q. Mao, Z. Wang, W. Yang, and S. Wang, ``Learned Image Compression for Both Humans and Machines via Dynamic Adaptation,'' in \emph{ICIP}, 2024.

\bibitem{mentzer2019}
F. Mentzer et al., ``Practical Full Resolution Learned Lossless Image Compression,'' in \emph{CVPR}, 2019.

\bibitem{bai2024dlpr}
Y. Bai, X. Liu, K. Wang, X. Ji, X. Wu, and W. Gao, ``Deep Lossy Plus Residual Coding for Lossless and Near-Lossless Image Compression,'' \emph{IEEE TPAMI}, 2024.

\bibitem{graziosi2020}
D. Graziosi et al., ``An Overview of Ongoing Point Cloud Compression Standardization Activities: Video-Based (V-PCC) and Geometry-Based (G-PCC),'' \emph{APSIPA Trans. Signal Inf. Process.}, 2020.

\bibitem{que2021}
Z. Que, G. Lu, and D. Xu, ``VoxelContext-Net: An Octree Based Framework for Point Cloud Compression,'' in \emph{CVPR}, 2021.

\bibitem{cui2023octformer}
M. Cui, J. Long, M. Feng, B. Li, and K. Huang, ``OctFormer: Efficient Octree-Based Transformer for Point Cloud Compression with Local Enhancement,'' in \emph{AAAI}, 2023.

\bibitem{nguyen2021}
D. T. Nguyen, M. Quach, G. Valenzise, and P. Duhamel, ``Learning-Based Lossless Compression of 3D Point Cloud Geometry,'' in \emph{ICASSP}, 2021.

\bibitem{wang2021pcgcv2}
J. Wang, D. Ding, Z. Li, and Z. Ma, ``Multiscale Point Cloud Geometry Compression,'' in \emph{Data Compression Conference (DCC)}, 2021.

\bibitem{sheng2022deeppcac}
X. Sheng, L. Li, D. Liu, Z. Xiong, Z. Li, and F. Wu, ``Deep-PCAC: An End-to-End Deep Lossy Compression Framework for Point Cloud Attributes,'' \emph{IEEE Trans. Multimedia}, 2022.

\bibitem{fu2026dald}
C. Fu, G. Li, W. Gao, S. Wang, Z. Li, and S. Liu, ``DALD-PCAC: Density-Adaptive Learning Descriptor for Point Cloud Lossless Attribute Compression,'' \emph{ACM TOMM}, 2026.

\bibitem{wu2024geoprior}
X. Wu, P. Zhang, M. Wang, P. Chen, S. Wang, and S. Kwong, ``Geometric Prior Based Deep Human Point Cloud Geometry Compression,'' \emph{IEEE Trans. Circuits Syst. Video Technol.}, 2024.

\bibitem{polygen}
C. Nash, Y. Ganin, S. M. A. Eslami, and P. W. Battaglia, ``PolyGen: An Autoregressive Generative Model of 3D Meshes,'' in \emph{ICML}, 2020.

\bibitem{meshgpt}
Y. Siddiqui et al., ``MeshGPT: Generating Triangle Meshes with Decoder-Only Transformers,'' in \emph{CVPR}, 2024.

\bibitem{meshxl}
S. Chen et al., ``MeshXL: Neural Coordinate Field for Generative 3D Foundation Models,'' in \emph{NeurIPS}, 2024.

\bibitem{ngf2024}
V. Sivaram, R. Ramamoorthi, and T.-M. Li, ``Neural Geometry Fields for Meshes,'' in \emph{ACM SIGGRAPH}, 2024.

\bibitem{ndf2025}
S. K. Pentapati, G. Phillips, and A. C. Bovik, ``Mesh Compression with Quantized Neural Displacement Fields,'' \emph{Computer Graphics Forum}, 2025.

\bibitem{chen2023npm}
Y.-C. Chen, V. G. Kim, N. Aigerman, and A. Jacobson, ``Neural Progressive Meshes,'' in \emph{ACM SIGGRAPH Conference Papers}, 2023.

\bibitem{witten1987}
I. H. Witten, R. M. Neal, and J. G. Cleary, ``Arithmetic Coding for Data Compression,'' \emph{Commun. ACM}, 1987.

\bibitem{vaswani2017}
A. Vaswani et al., ``Attention Is All You Need,'' in \emph{NeurIPS}, 2017.

\bibitem{liu2021}
Z. Liu et al., ``Swin Transformer: Hierarchical Vision Transformer Using Shifted Windows,'' in \emph{ICCV}, 2021.

\bibitem{zhao2021pt}
H. Zhao, L. Jiang, J. Jia, P. H. S. Torr, and V. Koltun, ``Point Transformer,'' in \emph{ICCV}, 2021.

\bibitem{velickovic2018gat}
P. Veli\v{c}kovi\'{c} et al., ``Graph Attention Networks,'' in \emph{ICLR}, 2018.

\bibitem{lee2019set}
J. Lee, Y. Lee, J. Kim, A. Kosiorek, S. Choi, and Y. W. Teh, ``Set Transformer: A Framework for Attention-based Permutation-Invariant Neural Networks,'' in \emph{ICML}, 2019.

\bibitem{jaegle2021perceiver}
A. Jaegle, F. Gimeno, A. Brock, O. Vinyals, A. Zisserman, and J. Carreira, ``Perceiver: General Perception with Iterative Attention,'' in \emph{ICML}, 2021.

\bibitem{meshopt}
A. Kapoulkine, ``meshoptimizer: Mesh optimization library,'' v0.25, \url{https://github.com/zeux/meshoptimizer}.

\bibitem{uvgvm}
G. Gautier, A. Mercat, L. Fr\'eneau, M. Pitk\"anen, and J. Vanne, ``UVG-VPC: Voxelized Point Cloud Dataset for Visual Volumetric Video-based Coding,'' in \emph{QoMEX}, 2023.

\bibitem{owlii}
Y. Xu, Y. Lu, and Z. Wen, ``Owlii Dynamic Human Mesh Sequence Dataset,'' ISO/IEC JTC1/SC29/WG11 m41658, 120th MPEG Meeting, Macau, 2017.

\bibitem{pytorch3d}
N. Ravi et al., ``PyTorch3D,'' arXiv:2007.08501, 2020.

\bibitem{pyg}
M. Fey and J. E. Lenssen, ``Fast Graph Representation Learning with PyTorch Geometric,'' in \emph{ICLR Workshop}, 2019.

\bibitem{open3dgc}
K. Mamou, \emph{Open3DGC}, \url{https://github.com/rbsheth/Open3DGC}.

\end{thebibliography}
\end{document}